\documentclass[conference]{IEEEtran}

\usepackage{times}

\usepackage[numbers]{natbib}
\usepackage{multicol}
\usepackage{hyperref}
\hypersetup{
    bookmarks=true,         % show bookmarks bar?
    unicode=false,          % non-Latin characters in Acrobat’s bookmarks
    pdftoolbar=true,        % show Acrobat’s toolbar?
    pdfmenubar=true,        % show Acrobat’s menu?
    pdffitwindow=false,     % window fit to page when opened
    pdfstartview={FitH},    % fits the width of the page to the window
    pdfnewwindow=true,      % links in new PDF window
    colorlinks=true,       % false: boxed links; true: colored links
    linkcolor=magenta,          % color of internal links (change box color with linkbordercolor)
    citecolor=red,        % color of links to bibliography
    filecolor=cyan,         % color of file links
    urlcolor=magenta        % color of external links
}
 
\usepackage{color, colortbl}
\definecolor{Gray}{gray}{0.95}
\usepackage{multirow}
\usepackage{ifthen}
\usepackage{graphicx}
\usepackage{svg}
\usepackage{caption}
\usepackage[labelformat=simple]{subcaption}
\usepackage{amsmath,amssymb,amsfonts}
\usepackage{algorithmic}
\usepackage{booktabs}       % professional-quality tables
\usepackage{graphicx}
\usepackage{textcomp}
\usepackage{xcolor}

\let\labelindent\relax % legacy IEEE template issue
\usepackage{enumitem}
\usepackage{microtype}
\usepackage{graphicx}
\usepackage{booktabs}
\usepackage{tcolorbox}

\definecolor{colorbox-color}{cmyk}{0.80, 0.13, 0.14, 0.04, 1.00}

\begin{document}

\title{Cherry-Picking with Reinforcement Learning: \\
Robust Dynamic Grasping in Unstable Conditions}

% You will get a Paper-ID when submitting a pdf file to the conference system
%\author{Author Names Omitted for Anonymous Review. Paper-ID 378}
% \author{Yunchu Zhang*, Liyiming Ke*, Abhay Deshpande, Abhishek Gupta, Siddhartha Srinivasa}

%\footnote{$*$: equal contribution}

%\author{\authorblockN{Michael Shell}
%\authorblockA{School of Electrical and\\Computer Engineering\\
%Georgia Institute of Technology\\
%Atlanta, Georgia 30332--0250\\
%Email: mshell@ece.gatech.edu}
%\and
%\authorblockA{Twentieth Century Fox\\
%\and
%\authorblockN{James Kirk\\ and Montgomery Scott}
%\authorblockA{Starfleet Academy\\
%San Francisco, California 96678-2391\\

% avoiding spaces at the end of the author's lines is not a problem with
% conference papers because we don't use \thanks or \IEEEmembership

% \author{Yunchu Zhang*, Liyiming Ke*, Abhay Deshpande, Abhishek Gupta, Siddhartha Srinivasa}
\author{\authorblockN{Yunchu Zhang\authorrefmark{1}\authorrefmark{2},
Liyiming Ke\authorrefmark{1}\authorrefmark{3},
Abhay Deshpande\authorrefmark{3}, 
Abhishek Gupta\authorrefmark{3} and
Siddhartha Srinivasa\authorrefmark{3}}
\authorblockA{\authorrefmark{1}\footnotesize{equal contribution}}
\authorblockA{\authorrefmark{2}The Robotics Institute, Carnegie Mellon University, Pittsburgh, USA}
\authorblockA{\authorrefmark{3}Paul G Allen School of Computer Science and Engineering, Seattle, USA}
}
\newcommand{\kay}[1]{\textcolor{cyan}{[kay: #1]}}
\newcommand{\yc}[1]{\textcolor{red}{[yunchu: #1]}}
\newcommand{\ssnote}[1]{\textcolor{red}{[sidd: #1]}}
\newcommand{\rsnote}[1]{\textcolor{orange}{[rosario: #1]}}
\maketitle

\IEEEpeerreviewmaketitle

\begin{abstract}

Grasping small objects surrounded by unstable or non-rigid material plays a crucial role in applications such as surgery, harvesting, construction, disaster recovery, and assisted feeding. This task is especially difficult when fine manipulation is required in the presence of sensor noise and perception errors; errors inevitably trigger dynamic motion,  which is challenging to model precisely. Circumventing the difficulty to build accurate models for contacts and dynamics, data-driven methods like reinforcement learning (RL) can optimize task performance via trial and error, reducing the need for accurate models of contacts and dynamics. Applying RL methods to real robots, however, has been hindered by factors such as prohibitively high sample complexity or the high training infrastructure cost for providing resets on hardware. This work presents CherryBot, an RL system that uses chopsticks for fine manipulation that surpasses human reactiveness for some dynamic grasping tasks. By integrating imprecise simulators, suboptimal demonstrations and external state estimation, we study how to make a real-world robot learning system sample efficient and general while reducing the human effort required for supervision. Our system shows continual improvement through 30 minutes of real-world interaction:  through reactive retry, it achieves an almost 100\% success rate on the demanding task of using chopsticks to grasp small objects swinging in the air. We demonstrate the reactiveness, robustness and generalizability of CherryBot to varying object shapes and dynamics (e.g., external disturbances like wind and human perturbations). Videos are available at \href{https://goodcherrybot.github.io/}{\color{red}{https://goodcherrybot.github.io/}}
\end{abstract}

\section{Introduction}

% The Problem
How can we automate the task of picking cherries from a tree branch that is blowing in the wind, causing the branch to shake and the cherries to tremble?
This scenario is an example of fine grasping \emph{without rigid-surface support},  and its challenges are two-fold. First, for fine manipulation of small objects, perception errors and sensor noise dominate, making it difficult to grasp the objects  precisely~\cite{cutkosky2012robotic, ke2021grasping}. Second, the problem is inherently dynamic since any contact with the object might set the entire scene into motion, which is complicated to model~\cite{mason1993dynamic, billard2019trends}. Similar challenges arise in our everyday interactions, from mundane tasks such as removing broken shells from gelatinous egg whites to surgical tasks that detach clots from deformable organs. Given the ubiquitous nature of these tasks, developing robotic solutions to automate them holds immense practical and economic value. 

For a predetermined, specific task, it is possible to invest in dedicated hardware~\cite{marohn2004davinci, yuan2017gelsight}, specialized tools~\cite{bhattacharjee2019towards, li2019vacuum, zeng2022robotic}, and elaborately designed systems~\cite{hwang2022automating, lynch1999dynamic} to solve these challenges. However, this research investigates a more universal solution: assuming that fine manipulation is required, inaccuracy is unavoidable and real-time reaction is necessary, can we enable dynamic fine grasping without stable support? An ideal agent should be:

% WOW Figure
\begin{figure}[!t]
\centering
% Use linewidth to adjust width
% Use textheight(the total height of the column of text) to adjust height
%\includegraphics[width=\linewidth]{media/focus-water.png}
  \begin{subfigure}[t]{.475\linewidth}
    \centering
    \includegraphics[width=\linewidth]{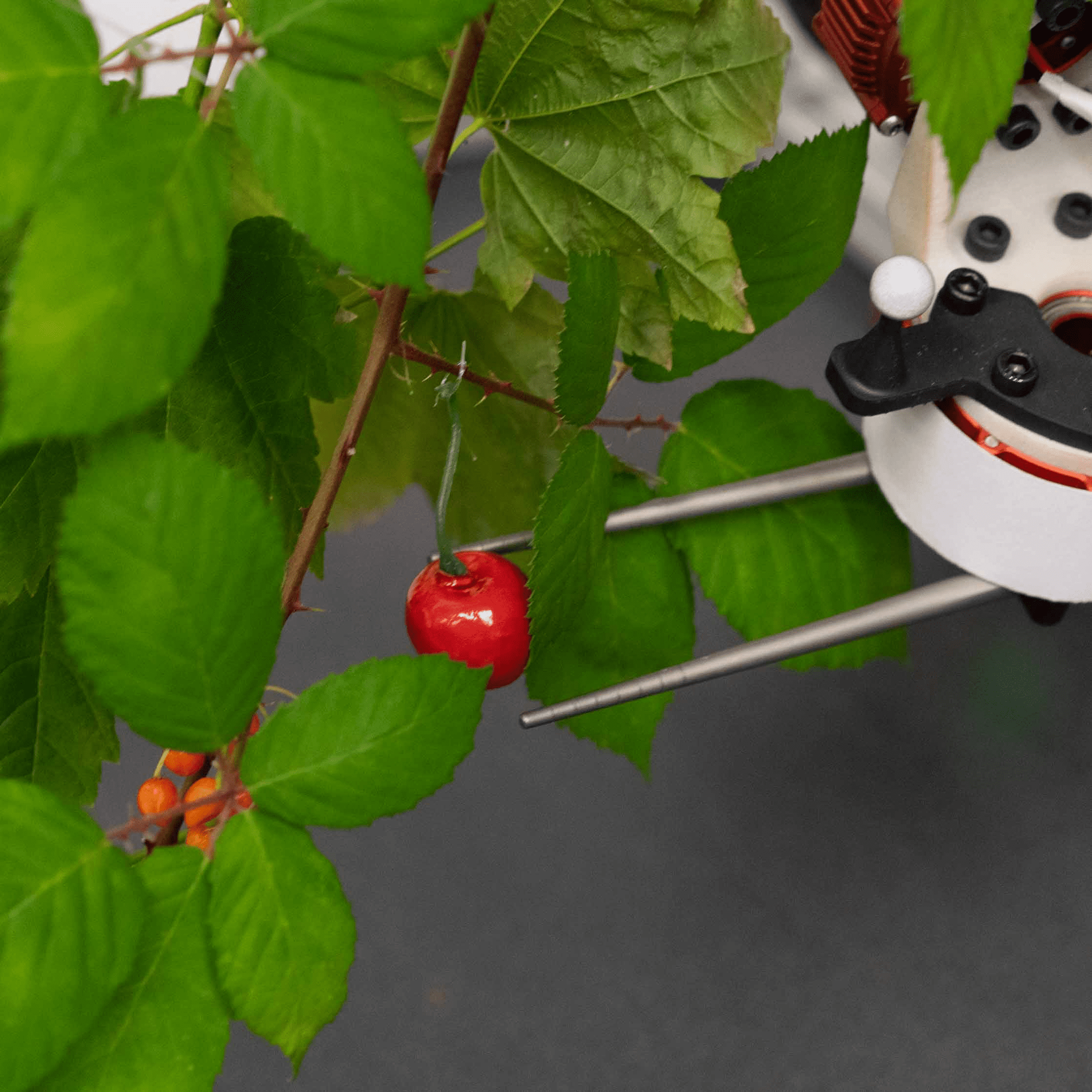}
  \end{subfigure}
  \hfill
  \begin{subfigure}[t]{.475\linewidth}
    \centering
    \includegraphics[width=\linewidth]{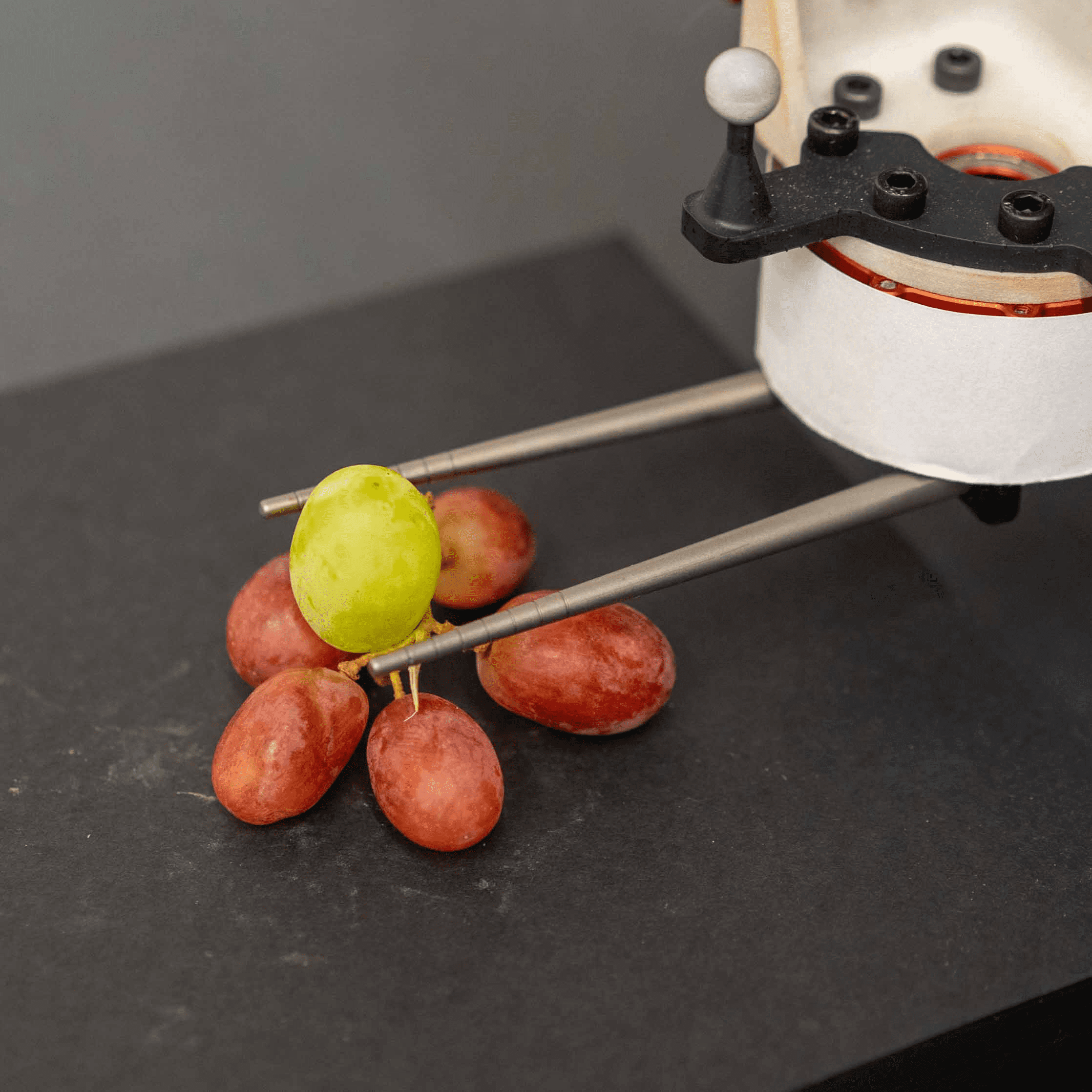}
  \end{subfigure}

  \vskip\baselineskip

  \begin{subfigure}[t]{.475\linewidth}
    \centering
    \includegraphics[width=\linewidth]{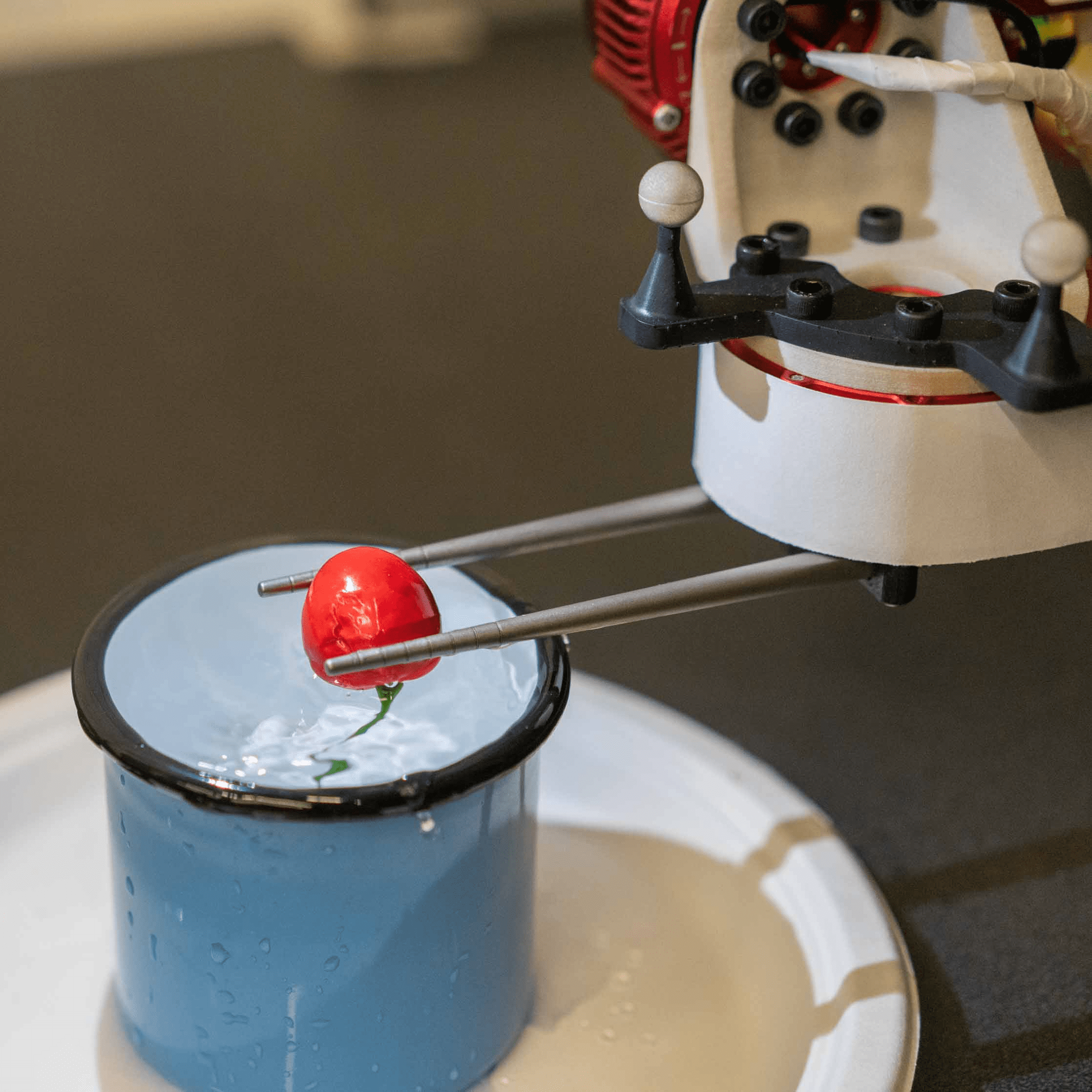}
  \end{subfigure}
  \hfill
  \begin{subfigure}[t]{.475\linewidth}
    \centering
    \includegraphics[width=\linewidth]{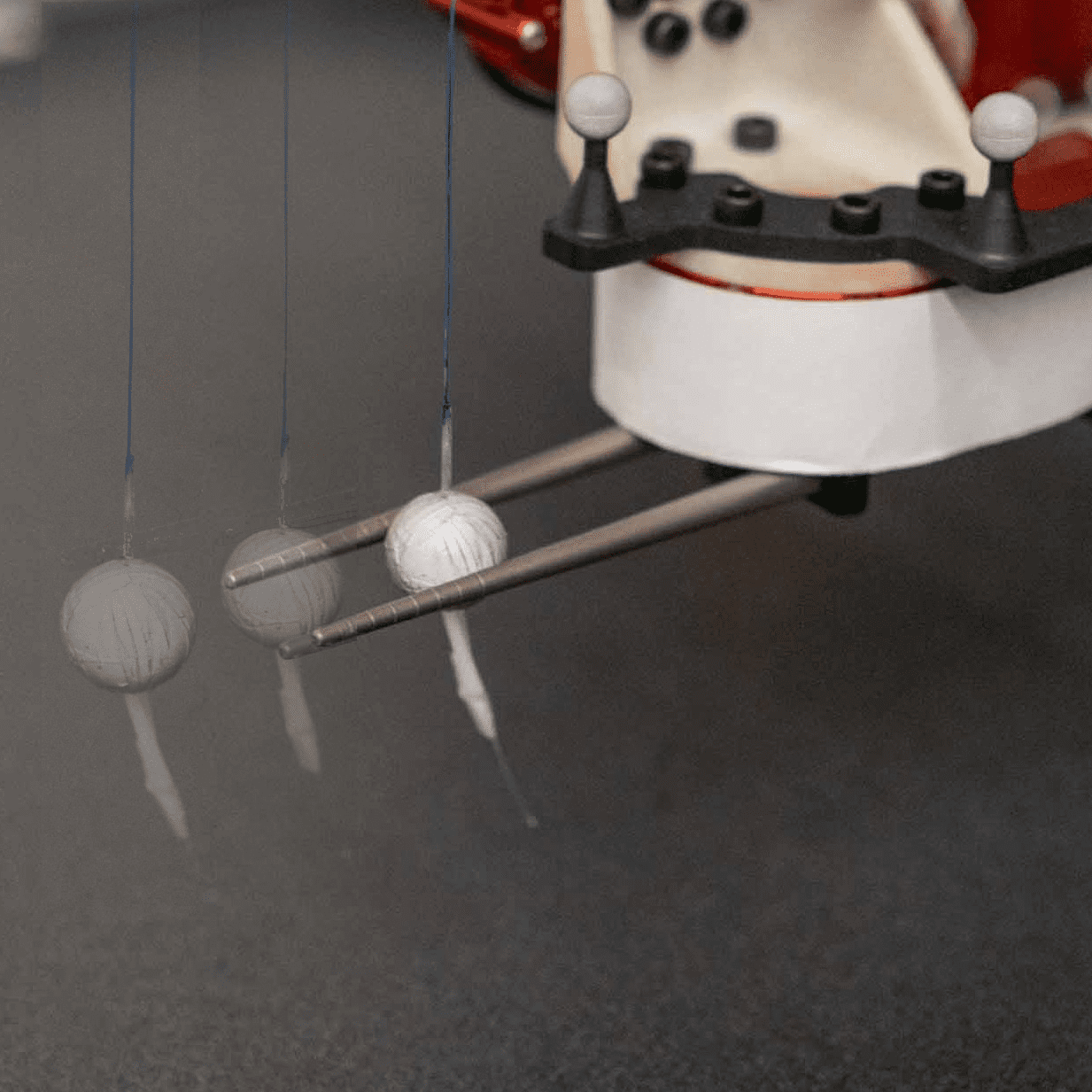}
  \end{subfigure}

\caption{\footnotesize{The CherryBot system generalizes to various scenarios of dynamic fine manipulation: blowing wind, sliding grape cluster, moving water, and swinging string.}}
\label{fig:best}
\vspace{-1em}
\end{figure}

\begin{itemize}
    \item \textbf{Precise} enough to increase the likelihood of task success. 
    \item \textbf{Robust} to perception errors and sensor noises that are likely to arise in the fine manipulation domain.
    \item \textbf{Reactive} to hard-to-model dynamic scenarios, external perturbations, and changes caused by its own movements.
    \item \textbf{Generalizable} to objects with different sizes, shapes and textures.
\end{itemize}

\begin{figure*}[t!]
\centering
\includegraphics[width=\textwidth]{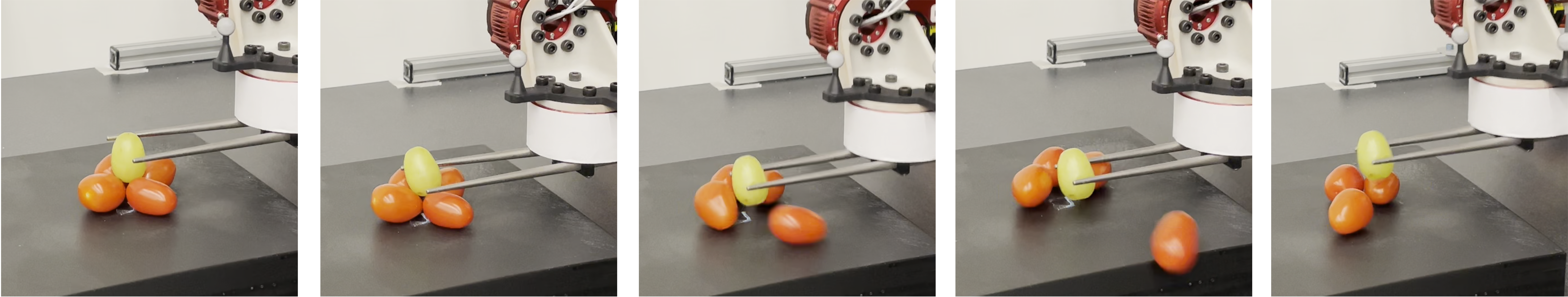}
\caption{\footnotesize{The dynamic grasping task is challenging: any contact with an object might set it into motion, which is difficult to model. This challenge is exacerbated when fine manipulation is required, especially in the presence of sensor noise and perception errors. }}
\label{fig:series}
\vspace{-1.2em}
\end{figure*}

% why chopsticks
To address these challenges, we build a test bed using generic hardware that includes a robot arm and chopsticks for fine manipulation. The design of chopsticks is simple yet versatile and has been widely adopted in surgery~\cite{sakurai2016thin, joseph2010chopstick}, meal assistance~\cite{yamazaki2012autonomous} and micro-manipulation~\cite{ke2021grasping, ramadan2009developmental, ke2020telemanipulation}. We note that the thin rods of chopsticks make fine manipulation more difficult and that our assembled hardware has sensing and actuation inaccuracies. However, the insights drawn from our accessible setup could be easily transferred to other platforms or tools that are built to operate with higher precision~\cite{marohn2004davinci, mason2011generality, chang2007pincer}.

% why RL
Prior work constructed analytic models~\cite{mordatch2012discovery,kumar2016optimal,hogan2020reactive} or motion primitives~\cite{schaal2006dynamic} for manipulation tasks with rich contacts. Instead, we choose reinforcement learning (RL) to circumvent some of the complexity of building accurate models for contacts, dynamics, different objects, or external disturbances. %RL is a data-driven method that has the potential to improve a policy through trial and error and has shown potential in generalizing across objects and contacts.
Despite their impressive potential for generalizability~\cite{kalashnikov2018qt}, applications of RL remain limited for real robots due to, for example, sample efficiency~\cite{zhu2019dexterous} and the costs of resetting~\cite{zhu20ingredients}. Though careful system design has enabled successfully deployed RL systems to learn on locomotion and dexterous manipulation tasks~\cite{peng2020learning, zhu2019dexterous}, the characteristics of our problem, i.e., fine manipulation with precise contact and hard-to-model dynamics, raise additional challenges in robustness and reactivity. 

% Our key insight and contribution
It is tempting to bypass modeling entirely and directly deploy a model-free RL algorithm for training in the real world.  While this approach may eventually learn, it often necessitates tremendous human effort for supervision and resets and is likely to be too inefficient. To make the training more practical, we propose CherryBot, an RL system that combines pre-training in an imprecise simulation with fine-tuning in the real world. With an external state estimation module, CherryBot can be deployed across various scenarios and fulfills the aforementioned requirements:
\begin{itemize}
\item \textbf{Efficiency:} To enhance sample efficiency for real-world manipulation, we leverage imperfect information readily available to most robots, such as an inaccurate simulator and a heuristic-based baseline policy. %This bootstrap approach surprisingly accelerates RL training.
\item \textbf{Precision and Robustness:} We introduce a challenging task that involves a single \emph{hard-to-grasp} object with \emph{diverse dynamics} for real-world fine-tuning. While this intensifies the difficulty of the task and training, it minimizes the need for human intervention and promotes the learning of robust policies that are resilient to disturbances.
\item \textbf{Reactiveness:} We carefully design the action space to strike a balance between tractable learning (low-frequency control, slow response, short horizon) and responsiveness (high-frequency control, fast response, long-horizon reasoning). This design optimizes the system's ability to react swiftly while ensuring effective learning.
\item \textbf{Generalizable}: Our system supports the plug-and-play integration of an external state estimation module even if it introduces certain inaccuracies, allowing deployment across various downstream tasks.
\end{itemize}

%Contribution 
Our work contributes a system that, given only 30 minutes of interaction in the real world, achieves superhuman reactiveness on a dynamic, high-precision task: using chopsticks to grasp a slippery ball swinging in the air. We demonstrate the effectiveness of our system in a variety of evaluation conditions: operating under dynamic disturbances by human or environmental factors, varying perception noise, and changing object shapes and sizes, for which it outperforms a heuristic-based controller that requires hours of tuning. We conduct extensive ablations in a simulator and the real world to provide empirical evidence about how our design choices affect sample efficiency for deploying RL systems in the real world and to verify the robustness of our proposal.

%%%%%%%%%%%%%%%%%%%%%%%%%%%%%%%%
%%%%%%%%%%%%%%%% ANYTHING BELOW IS RE-WRITE / SKETCH  %%%%%%%%%%%%%%%%
%%%%%%%%%%%%%%%%%%%%%%%%%%%%%%%%

\section{Related Works}

\textbf{Dynamic fine grasping.} From precondition grasping like DexNet~\cite{mahler2016dex} and TransporterNet~\cite{zeng2021transporter} to a closed-loop, vision-based controller like Qt-Opt~\cite{kalashnikov2018qt}, most prior works grasped palm-sized objects in a quasi-static table-top setting~\cite{calli2017yale}. In our work, objects are much smaller. Even if the agent intends to grasp across the center of the cherry, for example, the execution alone demands a sub-millimeter precision. Several previous works addressed fine grasping~\cite{ke2021grasping, hwang2022automating} but are limited to static scenarios with rigid surface support. In contrast, we examine a dynamics scene in which failed grasps can potentially move the object. Researchers in the field of dynamic manipulation have developed highly capable systems~\cite{hashimoto2001visuomotor}, although these systems come with the drawback of relying on expensive dedicated vision systems. In our approach, however, we prioritize the use of generic and accessible hardware, which is more susceptible to making mistakes. Consequently, it becomes crucial to develop a reactive agent that can effectively recover from such errors. We modify the common evaluation criteria for grasping to allow an agent that fails to get a firm grasp in one shot to continue the trial until it achieves a successful grasp, similar to how humans address dynamic grasping challenge~\cite{ke2020telemanipulation}. 

%\kay{TODO: cite Matt Mason. Compare our work with tossing bot - grasping and tossing versus catching, explain that it is impossible to build an accurate estimator through iteration of trials, like tossingbot and iterative residual policy did, because everytime when we miss the catch, we may have disturbed the system and changed what we wanted to estimate.} 

\textbf{Reinforcement learning.} Despite its wide successes in simulator and locomotion \cite{zhu2019dexterous,haarnoja2018learning,haarnoja2018soft}, RL has limited real world applications in manipulation. The few successes of applying RL to dexterous manipulation came at the cost of sample inefficiency, which has not made it a preferred choice over model-based control or a hand-designed controller. \cite{kalashnikov2018qt} depends on the large scale of a robot arm farm; ~\cite{wuthrich2020trifinger} requires scalable and dedicated hardware. Previous works focused on model-free RL while our work leverages practical assumptions, including inaccurate simulation~\cite{peng2020learning}, imperfect demonstration~\cite{rajeswaran2017learning}, and normalization techniques ~\cite{UTD, smith2022walk} to boost sample efficiency for RL and show it is practical on real robots and can achieve superhuman reactiveness. 

%\textbf{Learning from Demonstration} Imitation Learning has shown promise in grasping in ... [cite] and ... [cite], has even shown success on fine manipulation \cite{ke2021grasping}. Our task, however, is challenging even for human experts and the cost of collecting demonstration is non-trivial: even for a subspace of our task where the objects start with static pose, human experts have X success rate and it would take Y amount of time to collect data. Instead, we turn to learn from demonstrations collected by hand-writen controller but observe that we suffer from covariate shift. In contrast, our proposed method leverages demonstrations for sample efficiency but eventually surpassed the performance of the demonstrator and developed superhuman strategy. 

\textbf{Offline reinforcement learning (ORL).} The recently emerging field of ORL~\cite{lange2012batch,levine2020offline} has shown some potential for leveraging offline datasets for RL training~\cite{awr,fujimoto2018off,kidambi2020morel,kumar2020conservative,kostrikov2021offline, zhou2022real}. However, whether doing online fine-tuning on top of pre-trained ORL agents could keep improving the performance remains an open question~\cite{hong2022confidence}.

\textbf{Visual Servoing (VS).} VS controls a robot using real-time visual feedback~\cite{chaumette2006visual,hutchinson1996tutorial}. %There are two main types of VS schemes~\cite{sanderson1983adaptive}: position-based and image-based. %Image-based VS's controll low is based on the error between current and target features on the image space without the need of 3D pose estimation. 
It usually requires estimating the pose of the object in the Cartesian space and deriving a control law, such as PID control, directly in the 3D space~\cite{sanderson1983adaptive}. 
Due to the requirement of precision, it is natural to apply VS to fine manipulation tasks. We follow this protocol in the  design of our heuristic controller. However, it would require a tremendous amount of expert knowledge and gain-tuning to make the controller generalize across different scenarios or be robust to disturbance and noise. Our baseline controller took hours of gain-tuning but still fails due to disturbances, exemplifying how VS can be sensitive to unexpected dynamics. 

\section{Method}
\label{sec:method}

\begin{figure}[t!]
\centering
\includegraphics[width=\linewidth]{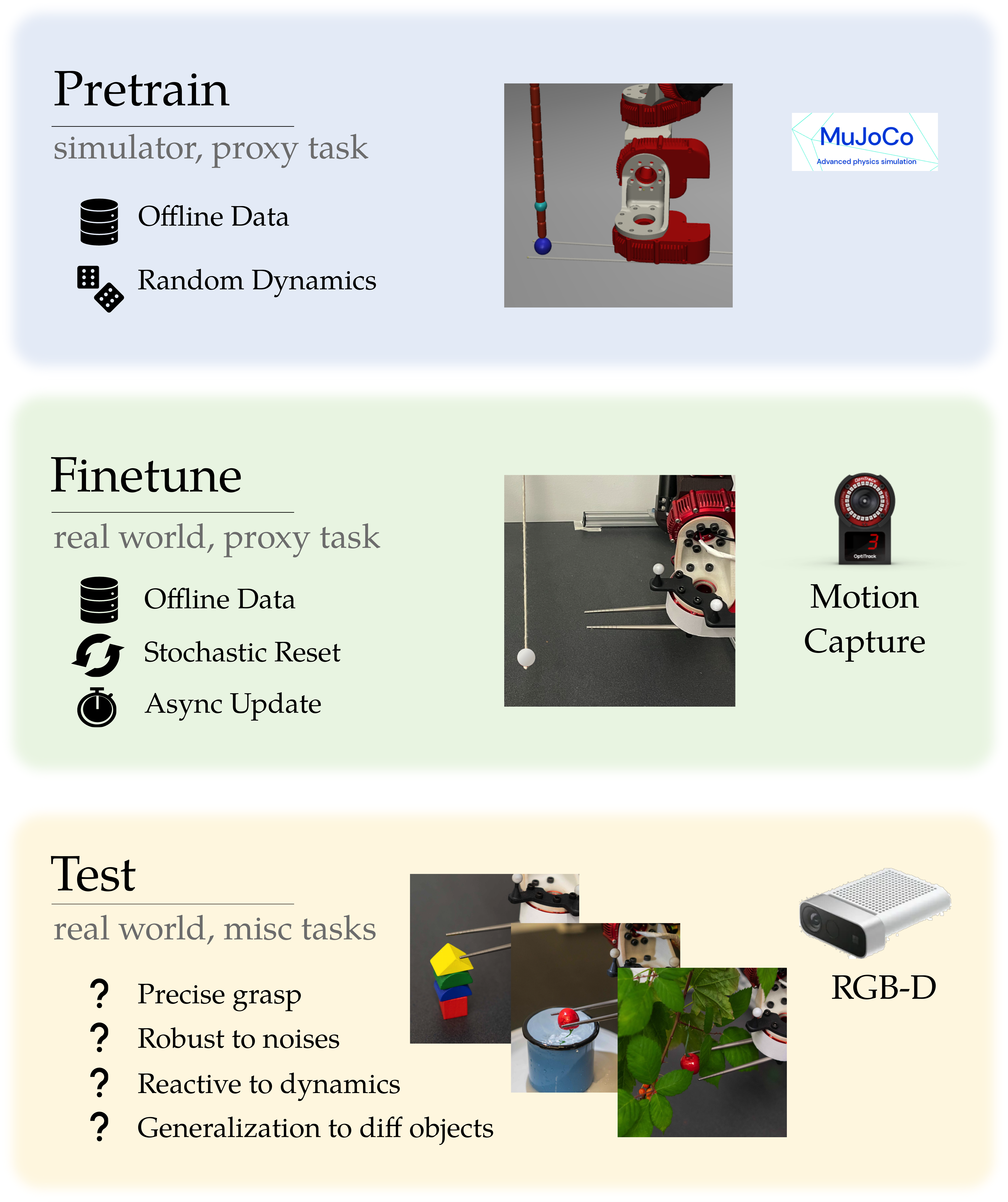}
\caption{\footnotesize{An overview of the learning framework. The system first pre-trains in simulation and then fine-tunes in the real world. We carefully design the training paradigm to ensure sample efficiency and learning robustness.}}
\label{fig:system}
\vspace{-1em}
\end{figure}

To address fine manipulation problems in dynamic scenes, we propose a comprehensive framework for efficient training of real-world RL and build a high-performance robotic system. We train a policy to enable a robot equipped with chopsticks to conduct dynamic fine-grasping tasks with non-rigid support. Following established practices in visual servoing~\cite{garrett2021integrated,zhang2022visually}, we incorporate a separate perception module, using estimated robot and environment states as input to the policy. Notably, our policy includes a \emph{hierarchical controller} (\texttt{Hier}) %consisting of a learned 20Hz policy that feeds to a low-level 1000Hz PID controller
to balance reactivity to dynamics and tractability of learning. At test time, we consider fine grasping of small objects (size $1\sim 8$ cm) with varying shapes and textures in a dynamic scene: objects might have unstable support and can be moving (e.g., initial velocity, or wind disturbance).

We present a system overview in Fig~\ref{fig:system}. Initially, we employ pre-training in an approximate simulator with domain randomization (\texttt{SimR}), leveraging sub-optimal demonstrations (\texttt{Demo}). Subsequently, we fine-tune the system in the real world on a task with a single object (\texttt{Proxy}) that presents dynamic challenges but simplifies perception and allows for easy and stochastic reset (\texttt{StoRe}). To further enhance training efficiency, we incorporate asynchronous updating (\texttt{Async}) and appropriately regularized off-policy RL (\texttt{LN}). During testing, the policy integrates an external perception module (\texttt{Vis}) to achieve generalization across different objects and scenes.

Our contribution lies in instantiating distinct system components to address the combined challenges of real-world RL and dynamic fine manipulation. We outline these challenges in Table~\ref{tab:pro_and_con} and note how each system component addresses them. To ascertain the impact of each design choice, we conduct ablation studies to measure sample efficiency in both simulation and the real world, where all studies are conducted using 5 consecutive random seeds (Refer to Appendix.~\ref{app:system} for details). While individual components may have been the subject of previous studies, the combination presented in our work is novel. Our system can grasp a diverse set of small objects in varying dynamic scenarios in the real world, demonstrating the robustness and generalizability to novel objects and disturbances, which could generalize to other tasks with dynamics or requiring fine manipulation.

In this section, we delve into the specifics of our design decisions, grouped into three categories: controller interface design, simulator pre-training and real-world fine-tuning. 

\begin{table}[htbp]
    \normalsize
    \setlength{\tabcolsep}{3pt}
    \centering
    \begin{tabular}{rcc|cccc}
    \toprule
             &\multicolumn{2}{c}{\textbf{RL challenges}} &\multicolumn{4}{c}{\textbf{Task challenges}} \\
    \midrule
    & Efficient & Reset & Robust & Reactive & Precise & General \\
    \midrule 
    \midrule
    \texttt{Hier} & \checkmark & & & & & \\
    \texttt{Demo}  & \checkmark & & & & & \\
    %\texttt{Sim} & \checkmark & & \checkmark & & & \\
    \texttt{SimR} & \checkmark & & \checkmark & & & \\
    \texttt{Proxy} & \checkmark & \checkmark & \checkmark & \checkmark & \checkmark & \\
    \texttt{StoRe}  & & \checkmark & \checkmark & & & \\
    \texttt{Async} & \checkmark & & & & & \\
    \texttt{LN} & \checkmark & & & & & \\
    \texttt{Vis} & \checkmark & & & & & \checkmark \\
    \bottomrule
    \end{tabular}
    \vspace{3mm}
    \caption{\footnotesize{Some challenges inherent to training RL algorithms for dynamic fine manipulation and the design decisions we make to address them.}}
    \label{tab:pro_and_con}
    \vspace{-1em}
\end{table}

% ===============================

\subsection{Hierarchical controller of mixed control frequency to balance tractability of learning and reactivity}

\begin{figure}[htbp]
\centering
  \begin{subfigure}{.24\textwidth}
    \centering
\includegraphics[width=\linewidth]{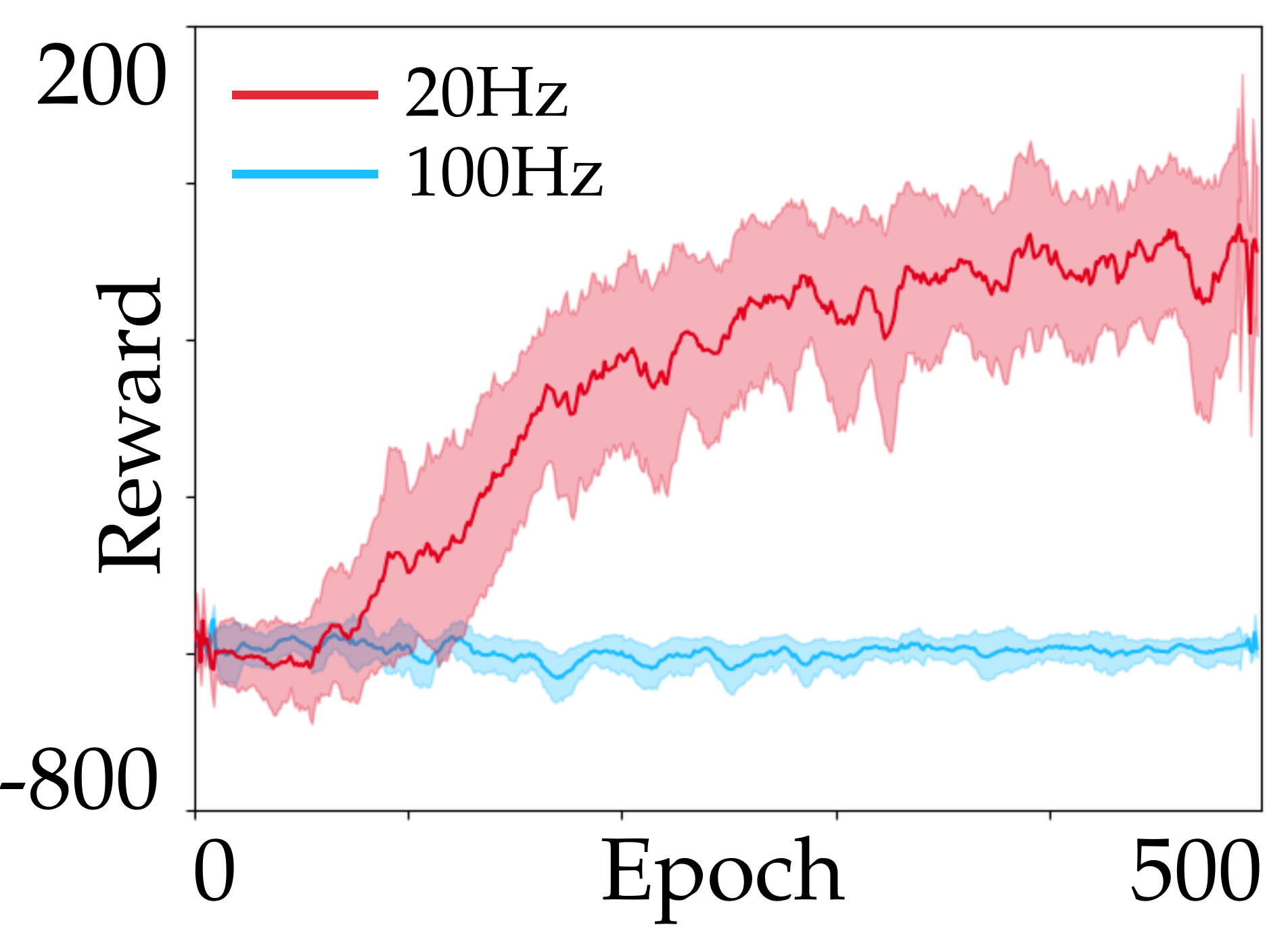}
    \caption{\footnotesize{Effect of control frequency}}  %A lower control frequency can improve sample efficiency.
    \label{fig:ablation-freq}
  \end{subfigure}
  \begin{subfigure}{.24\textwidth}
    \centering
\includegraphics[width=\linewidth]{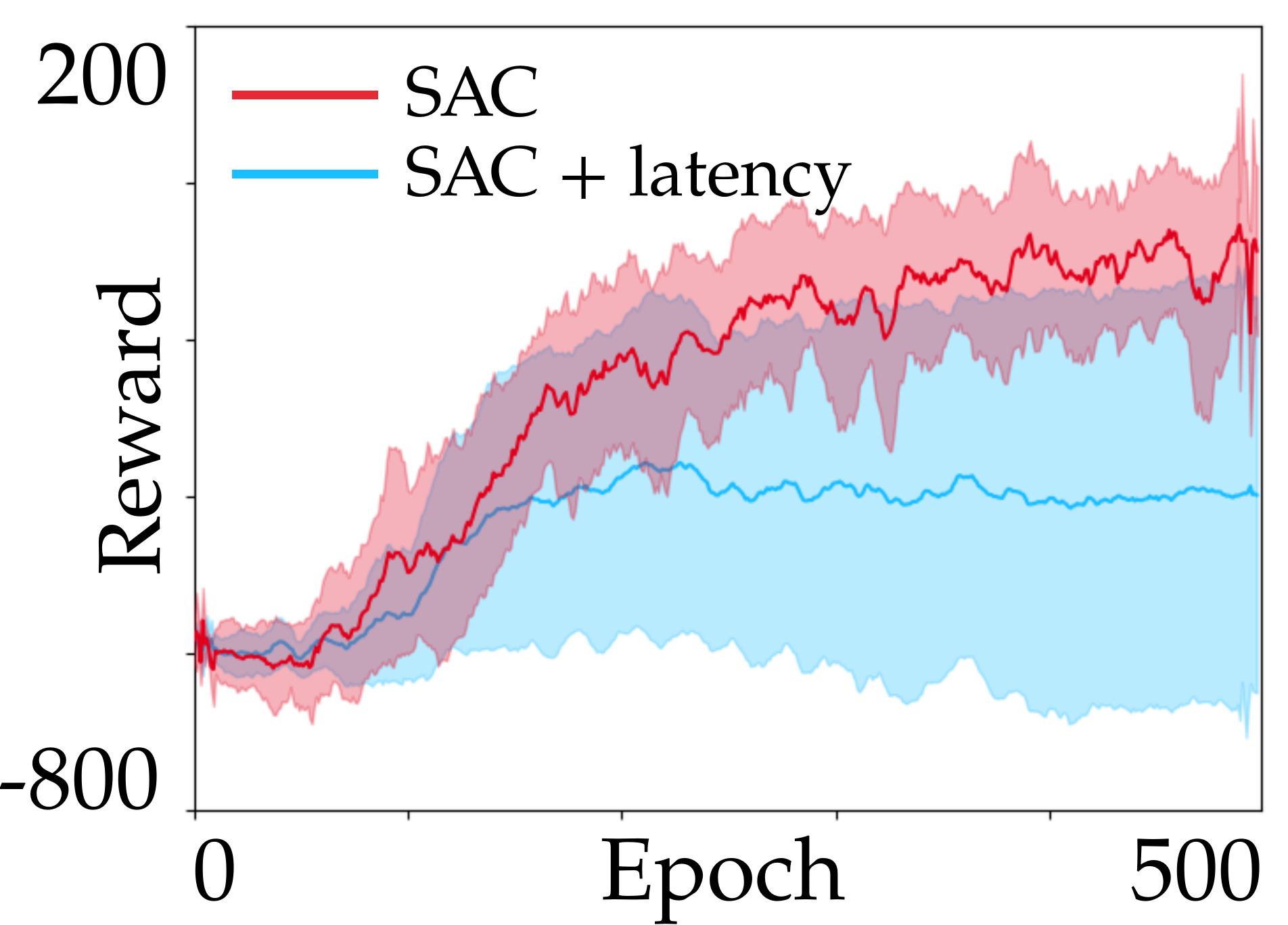}
    \caption{\footnotesize{Effect of latency ($< 0.005s$)}} %Latency (simulated) significantly harms the agents' ability to learn.
    \label{fig:ablation-latency}
  \end{subfigure}
\caption{\footnotesize{Analysis of using hierarchical mixed-frequency controllers in simulation. \textbf{Left:} Impact of different control frequencies on learnability. A hierarchical hybrid-frequency strategy helps balance learnability and policy reactivity. \textbf{Right:} Impact of (simulated) latency on performance. }}
\label{fig:method_hierarchical}
\vspace{-1em}
\end{figure}

Previous research on dynamic tasks has favored the development of 1000Hz controllers~\cite{wuthrich2020trifinger, chi2022iterative} and vision systems~\cite{okumura2011high}. However, as demonstrated in Fig~\ref{fig:ablation-freq}, high-frequency control can negatively impact task horizon and sample efficiency for robotic learning. Real-life RL manipulations commonly employ controllers operating at 5 $\sim$ 15 Hz for quasi-static tasks~\cite{zhu2019dexterous, zhou2022real}.

To address this challenge, we adopt a hierarchical controller framework, enabling the RL policy to function at a higher level with a moderate control frequency of 20 Hz, while interpolating commands to a high-frequency (1000Hz) controller. Our system demonstrates that such a controller is capable of solving dynamic tasks and, combined with coherent exploration, makes the learning of a reactive policy tractable.

Furthermore, in the real world, sensor readings suffer from latency and latency can adversely affect reactivity.
While~\cite{peng2018sim} demonstrated the advantages of latency randomization in simulation pre-training for a quasi-static task, our own experiments (Fig~\ref{fig:ablation-latency}) revealed that even a small simulated latency ($\leq 0.005$ s) can harm our task's reactivity to dynamics. Consequently, we made the deliberate decision to exclude simulating latency or latency randomization in simulator training.%, emphasizing the need for tailored considerations in dynamic tasks.

%This equates to the observation being lagged by one or more timesteps, which violates the Markov property and degrades performance. We experimentally verify this through ablation in Fig~\ref{fig:ablation-latency}, where we show that training using observations from in the recent past (that simulate latency) significantly impacts learning performance. We observe that the latency's impact becomes negligible when the control frequency is lower, which decreases the relative ratio of the length of the latency over the length of the control step.

%Therefore, we elect to use a hierarchical controller framework that allows the RL policy to operate at a higher level with a moderate control frequency (20Hz) and interpolate those commands to a low-level, high-frequency (1000Hz) controller. Our system shows that such a controller, with coherent exploration, can make it tractable to learn a reactive policy. 

\begin{tcolorbox}[colback=red!5!white,left=1mm,top=1mm,right=1mm,bottom=1mm]
\textbf{Insight:} Learning medium-frequency hybrid controllers can effectively balance policy reactivity with the tractability of learning. 
\end{tcolorbox}
% ===============================

\subsection{Leveraging practical information: Approximate  simulator \texttt{SimR} and sub-optimal offline data \texttt{Demo}}
\label{sec:pretrain}

An RL agent initialized from scratch in the real world makes random movements, frequently encounters safety constraints, and spends a significant portion of training time waiting to be reset. To counter these challenges, we turn to the paradigm of offline pre-training followed by real-world fine-tuning. Where should the data for pre-training actually come from? Since human data collection is expensive for our dynamic fine manipulation task, we instead rely on relatively cheap and abundant, but imperfect, sources of supervision, i.e., approximate simulators and data from heuristic controllers.

\begin{figure}[htbp]
\centering
\includegraphics[width=\linewidth]{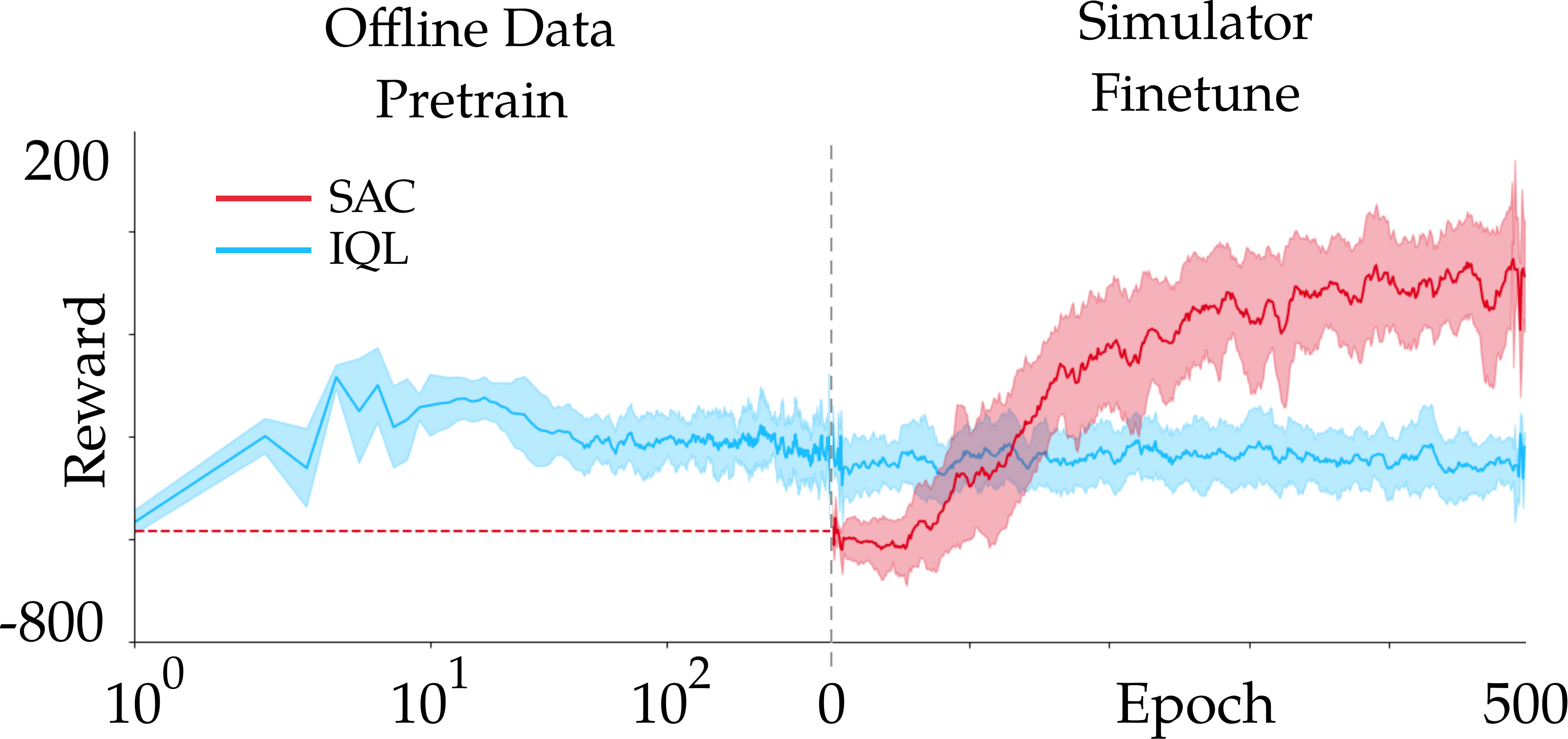}
    \caption{\footnotesize{Analysis of the choice of an off-policy RL algorithm that can learn from offline data while continuing to improve efficiently during fine-tuning.}}
    \label{fig:ablation-iql}
\vspace{-1em}
\end{figure}

%Since the above sources of data are imperfect,
\textbf{Choosing an appropriate off-policy algorithm.}
We first consider an appropriate RL algorithm that can efficiently learn from offline data, simulation samples and online fine-tuning.  Although various offline RL methods~\cite{kumar2020conservative, kostrikov2021offline} have been proposed to conduct first pre-training and then fine-tuning, we found that standard \emph{online} RL algorithms (such as soft actor critic (SAC) ~\cite{haarnoja2018soft}) are far more effective for fine-tuning than targeted offline RL methods, as Fig~\ref{fig:ablation-iql} shows. In our work, we used variants of soft actor critic~\cite{haarnoja2018soft} for both pre-training and fine-tuning. Our modifications are described below.

\textbf{Leveraging imperfect data from the simulation.} 
Simulation can provide an appealing source of information since sampling is cheap. However, our task depends heavily on precise contacts and dynamics, but our hardware has varying degrees of kinematic and dynamic errors, making it challenging to construct a precise simulator and making direct simulation-to-reality transfer difficult. 
%Instead, we simply use simulation for pre-training Q-functions and policies in our RL algorithm, relying on additional fine-tuning to correct for the bias in the simulation data.

\begin{figure}[!h]
\centering
  \begin{subfigure}{.47\linewidth}
    \centering
\includegraphics[width=\linewidth]{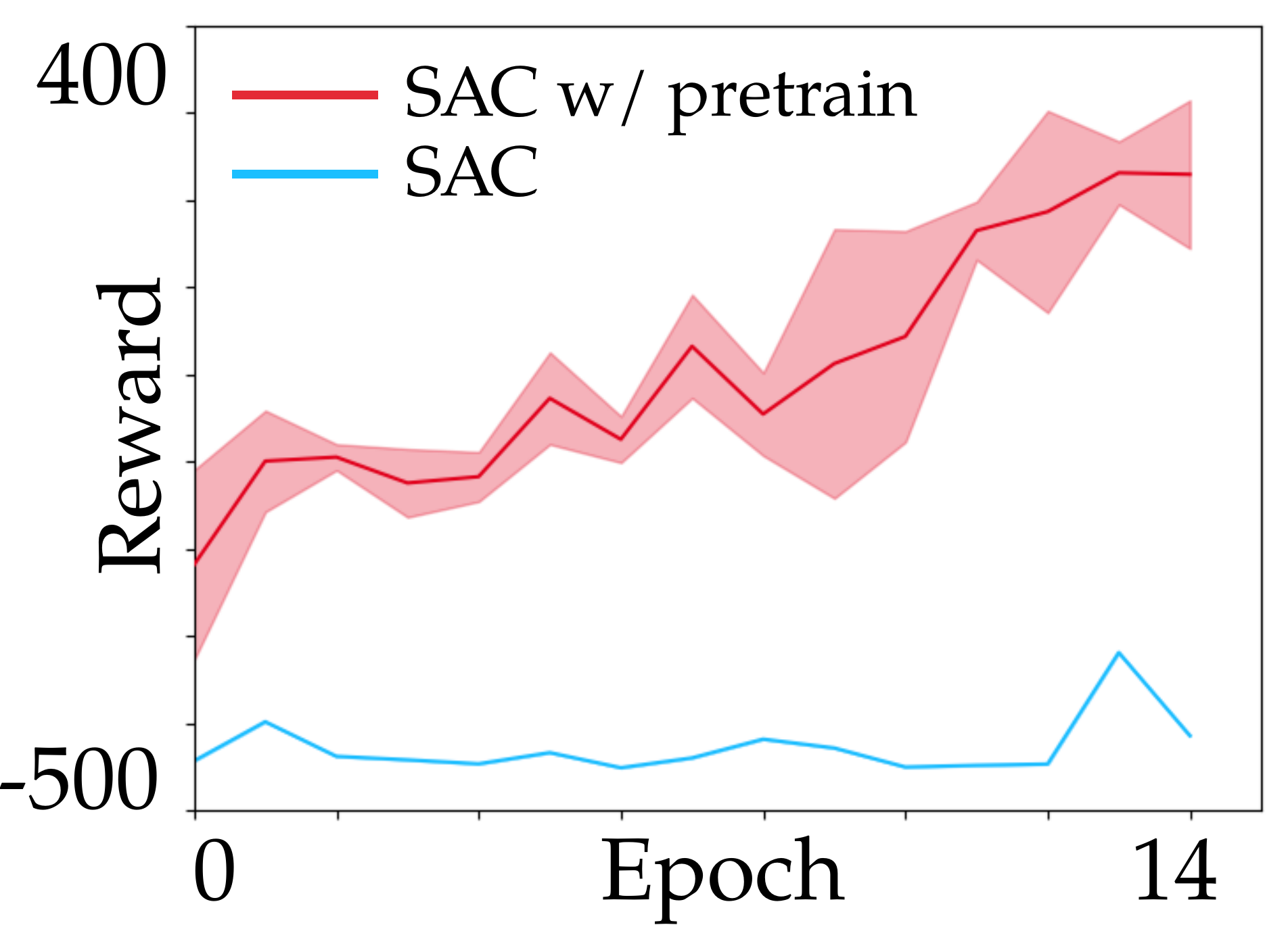}
    \caption{\footnotesize{Pre-train from simulation}}
    \label{fig:ablation-sim}
  \end{subfigure}
  \begin{subfigure}{.47\linewidth}
    \centering
\includegraphics[width=\linewidth]{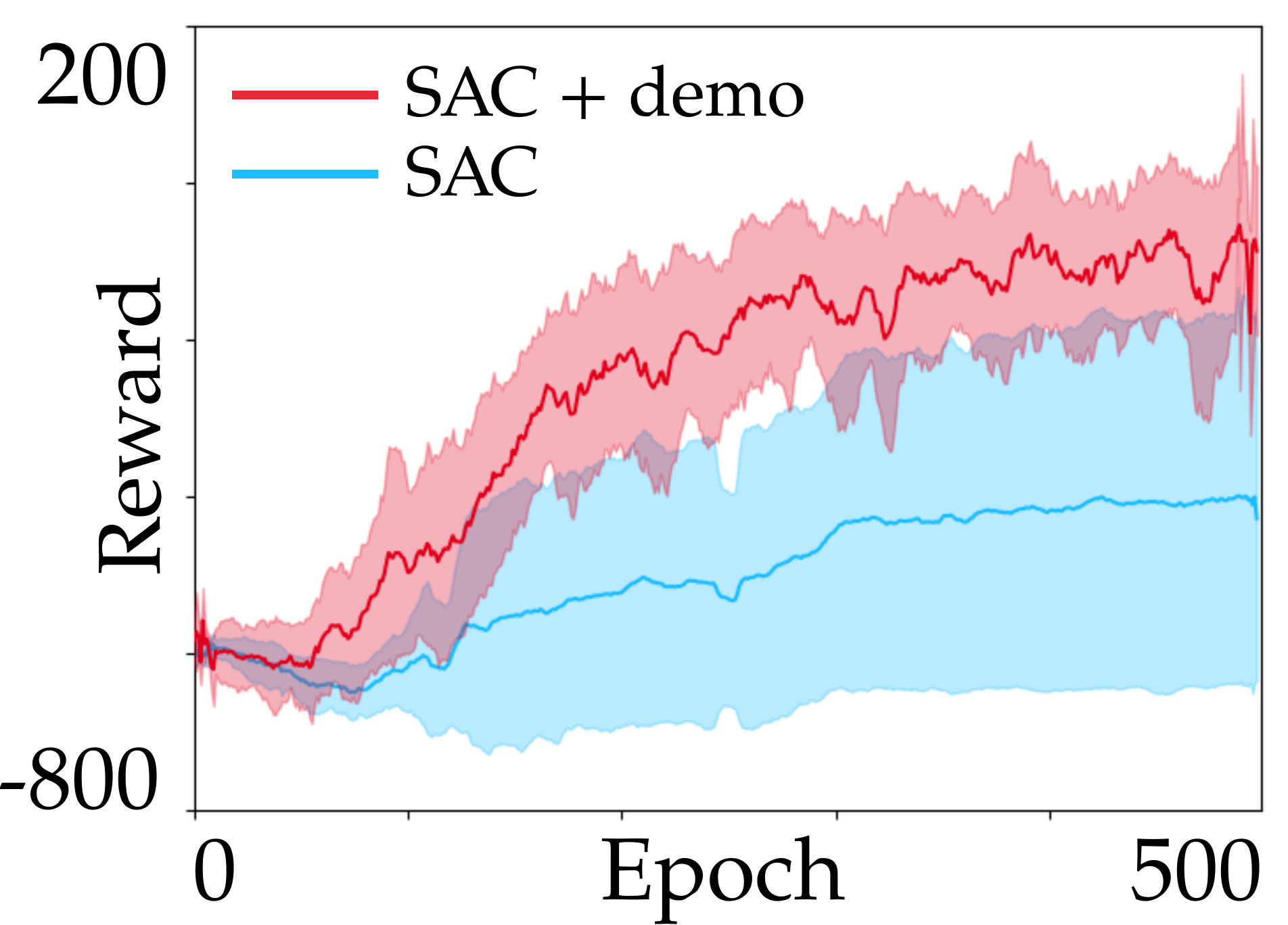}
    \caption{\footnotesize{Pre-train from heuristic controller}}
    \label{fig:ablation-demo}
  \end{subfigure}
\caption{\footnotesize{Analysis of the impact of pre-training for fine manipulation. \textbf{(Left)} Having pre-trained in simulation enabled real-world fine-tuning to make progress. \textbf{(Right)} Inclusion of offline data collected from a heuristic controller enabled simulator pre-training to succeed on the task in the simulator using many fewer samples. Both techniques significantly aid with learning efficiency.}}
\label{fig:method_pretrain}
\vspace{-1em}
\end{figure}

We constructed an approximate simulator, as described in Sec.~\ref{sec:system}. To best leverage such a simulation that is definitely mismatched with reality, we randomize the dynamics of the physical simulation~\cite{peng2018sim}, exposing the agent to a wide distribution of possible physics parameters. Unaware of the changing dynamics of the environment, the agent needs to develop a robust strategy that is conservative to variations in the world dynamics, making it more amenable to real-world fine-tuning. 
Specifically, we train a Q-function and policy networks with samples from our simulator and then use the networks to initialize the fine-tuning phase. Without pre-training, an RL agent could not improve its task performance even after hours of real-world training, as shown in Fig~\ref{fig:ablation-sim}.

\textbf{Leveraging sub-optimal offline data.} Beyond simulation data, it is useful to provide the agent access to real-world data with sufficient state-action coverage. While it may be expensive to solicit data from a human supervisor or to generate optimal demonstration, it is relatively easy to design a heuristic policy that is sub-optimal but can provide useful state action coverage.
% A common reason for the sample inefficiency in reinforcement learning training is that: the RL agent, having not encountered some of the good state spaces with high reward values, could not be incentivized to go to those state. We would like to provide the RL agent with real-world examples that cover the useful parts of the state action space. While it is challenging to hand-define a robust yet generally optimal controller in the real world, it is relatively easy to design a heuristic policy that is sub-optimal but can provide useful state action coverage. 
In this work, we design a simple heuristic controller that contains a state machine with closed-loop control (see Appendix~\ref{app:heuristic}). Though this controller has limited reactivity and is not robust to noise, we found that seeding the replay buffer of our RL agent with this sub-optimal data helps with both asymptotic performance and sample efficiency (Fig.~\ref{fig:ablation-demo}).

% This is perhaps caused by offline RL methods being overly conservative, making them difficult to aggressively improve via fine-tuning \cite{awr}. As we describe in the following sections, we obtain the data for pre-training from two sources - heuristic controllers in the real world and approximate simulations. We describe each pre-training from each of these data sources below.

\begin{tcolorbox}[colback=red!5!white,left=1mm,top=1mm,right=1mm,bottom=1mm]
\textbf{Insight:} Pre-training using standard off-policy RL methods with imperfect prior data from heuristic controllers and simulation can significantly help with sample efficiency for real-world fine-tuning.  
\end{tcolorbox}

\begin{figure}[!h]
\centering
\includegraphics[width=\linewidth]{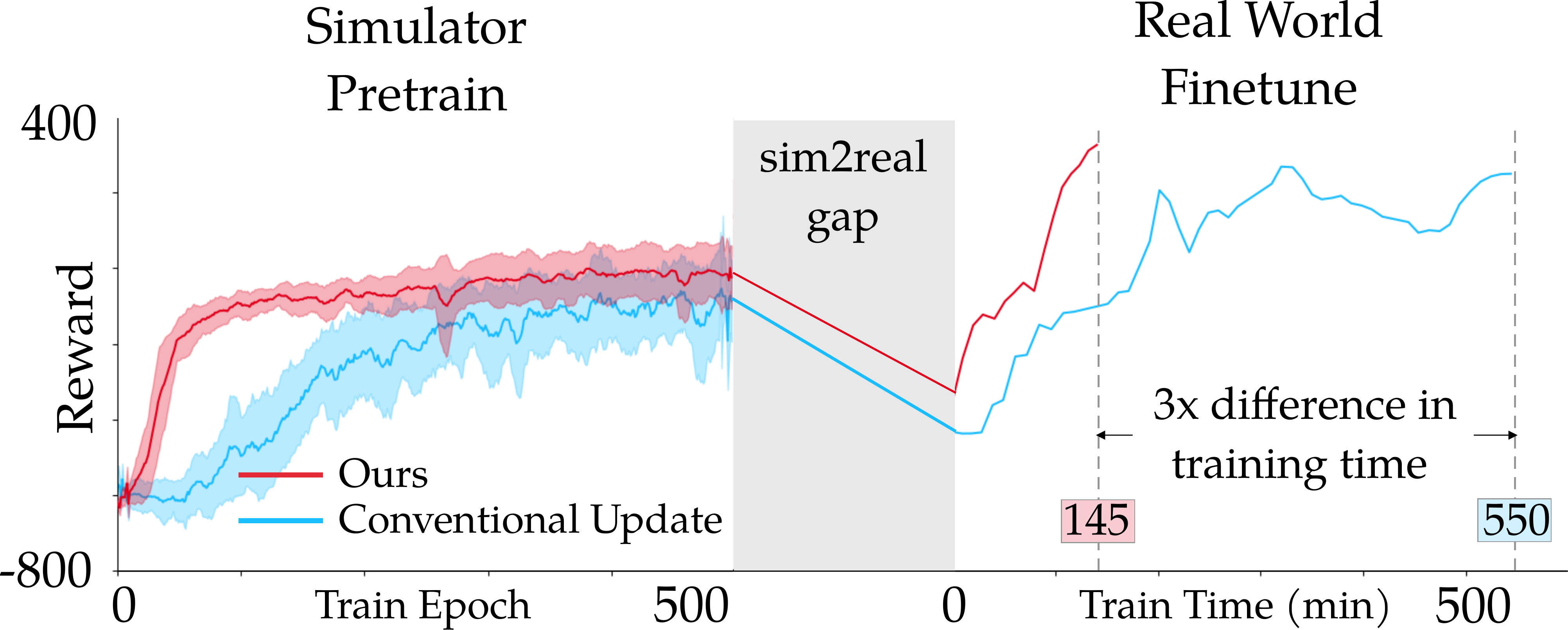}
    \caption{\footnotesize{We measure sample efficiency in simulation by training epochs but in the real world by training time. Combining the asynchronous update, high Update-To-Data (UTD) ratio and LayerNorm regularizer greatly improves the speed of training during the real-world finetuning stage.}}
    \label{fig:ablation-finetune}
\vspace{-1em}
\end{figure}
\subsection{Fine-tuning in the real world: Proxy task and       efficient reinforcement learning}
\label{sec:finetune}

\begin{figure*}[!t]
\centering
\begin{subfigure}[t]{.4\linewidth} %.5
\centering
\includegraphics[width=\linewidth]{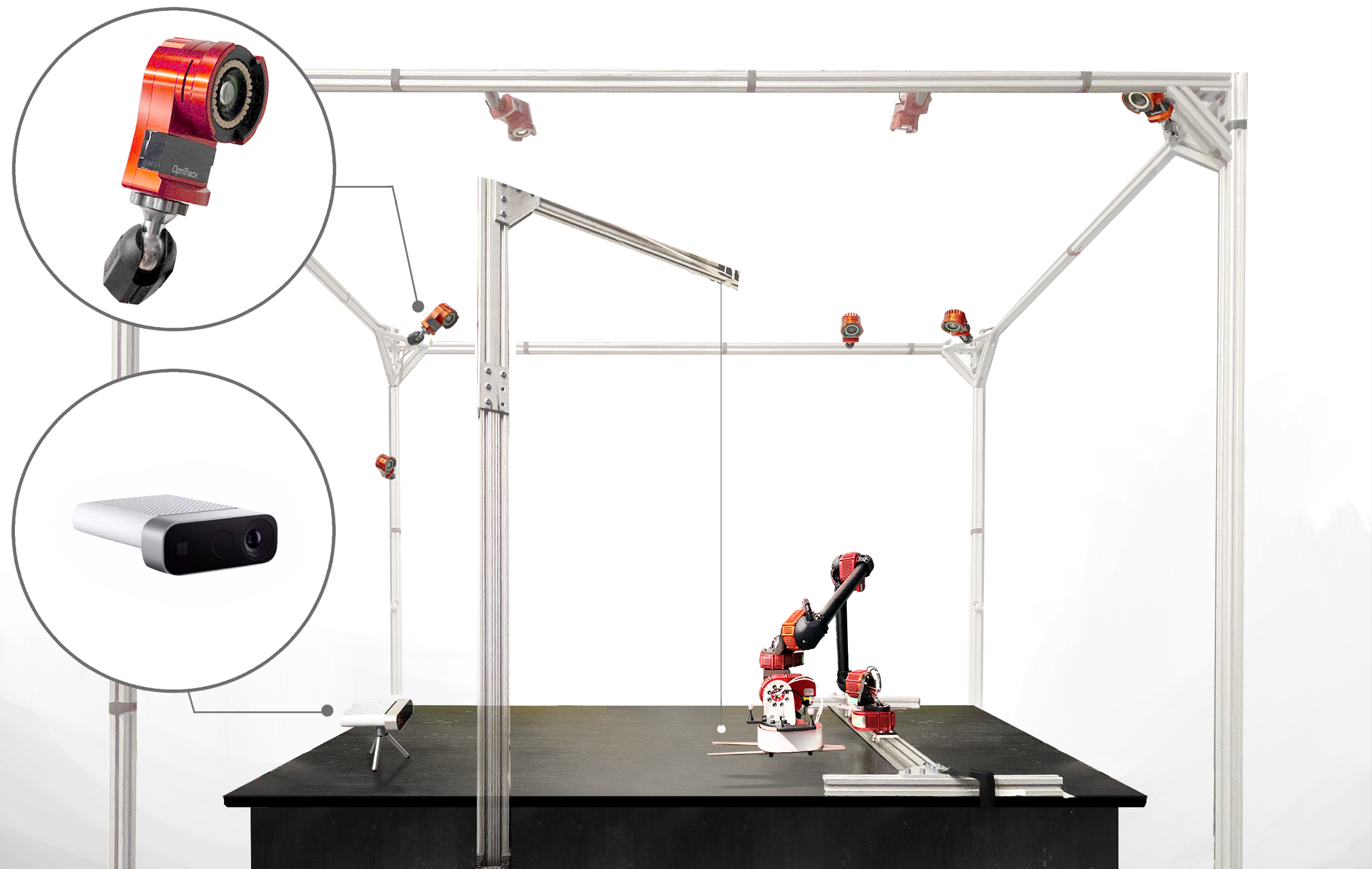}
\caption{Hardware overview.}
\label{fig:hardware_overview}
\end{subfigure}
\begin{subfigure}[t]{.12\linewidth} %.15
\centering
\includegraphics[width=\linewidth]{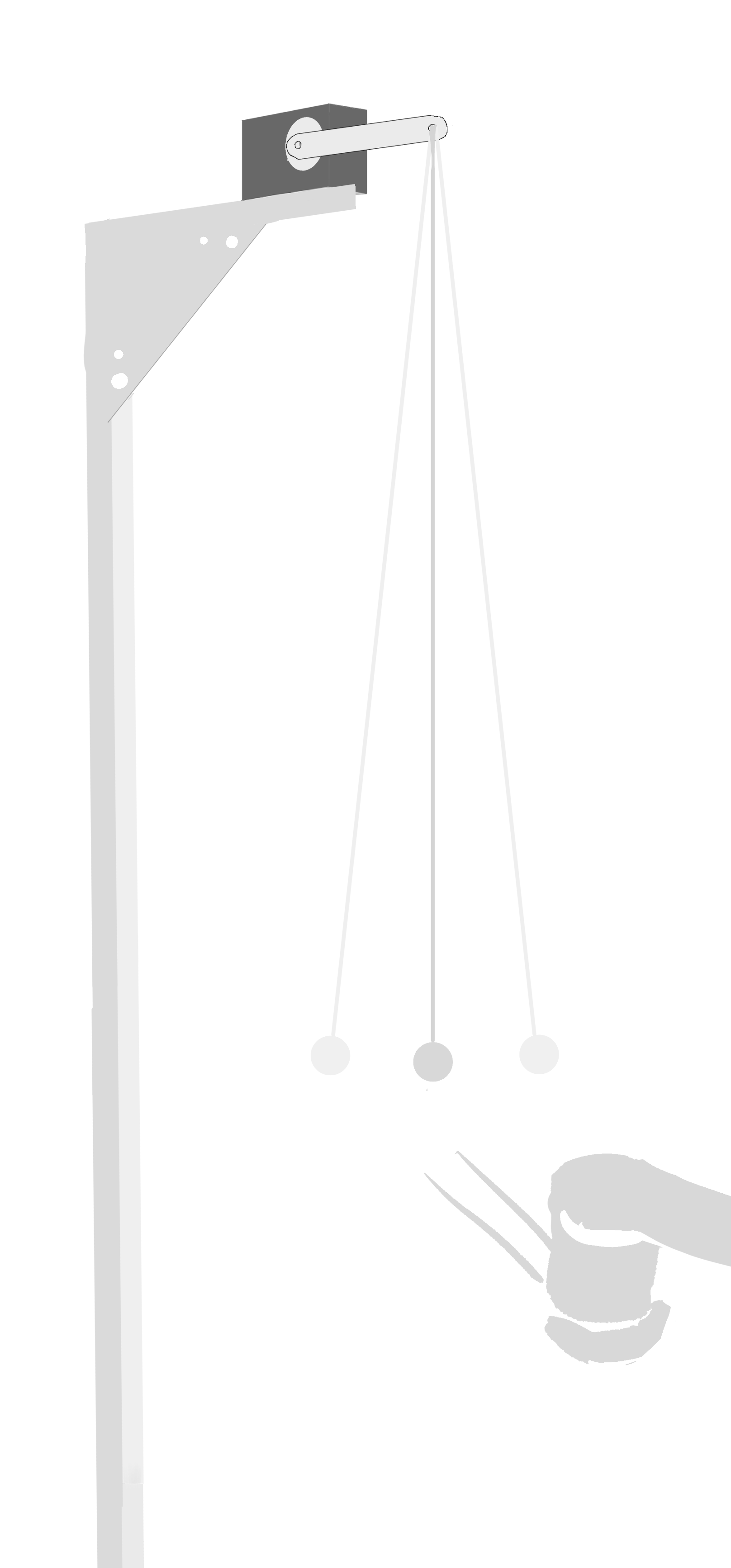}
\caption{Proxy task.}
\label{fig:proxy_task}
\end{subfigure}
\begin{subfigure}[t]{.21\linewidth} %.3
\centering
\includegraphics[width=\linewidth]{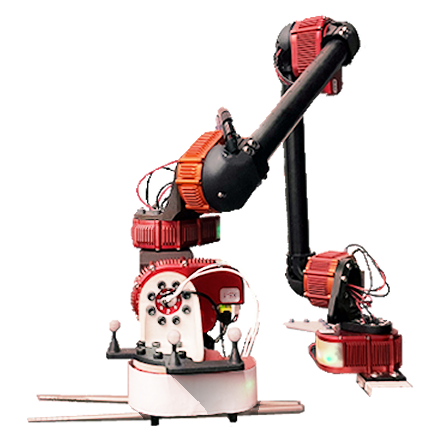}
\caption{Assembled robot.}
\label{fig:robot}
\end{subfigure}
\begin{subfigure}[t]{.22\linewidth} %.3
\centering
\includegraphics[width=\linewidth]{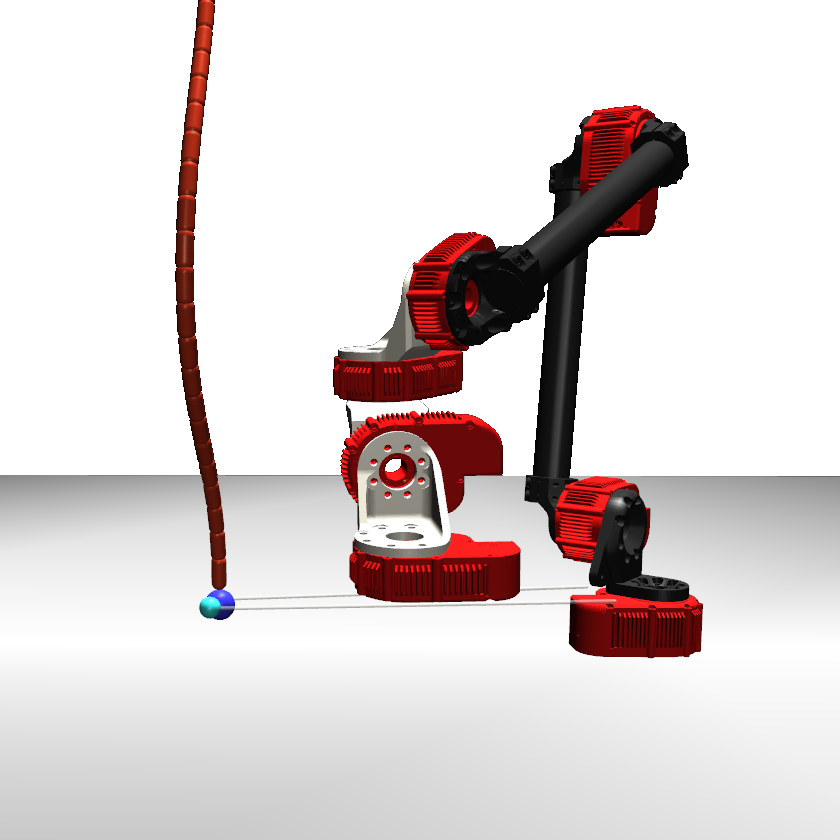}
\caption{MuJoCo simulation.}
\label{fig:robot_sim}
\end{subfigure}
\caption{\footnotesize{The CherryBot hardware system.}}
\vspace{-1em}
\end{figure*}

Although pre-training can significantly aid with data efficiency, it is unlikely to achieve optimal performance without further training in the real world. To enable practical fine-tuning in the real world, we focus on two questions: (1) what real-world training setup allows for both ease of training and robust learning, and (2) how do we make this process more efficient?

\textbf{Pragmatic fine-tuning via a single proxy task.} Conventional wisdom might suggest training the RL agent directly on a large variety of test scenarios. However, a real-world training setup would require a prohibitively large number of objects and scenarios to be instrumented, which can be laborious and expensive. Instead, we posit (1) to isolate the perception challenge and control challenge, and (2) first address the control challenge via a \texttt{Proxy} task that is appropriately dynamic and difficult, such that it can yield a policy that generalizes to a variety of dynamic scenarios without actually being explicitly exposed to them during training. We identify the criteria needed to set up a proxy task as follows: (1) \emph{representative} of the motion and strategies involved in a broader family of tasks, (2) \emph{reset-friendly}, or having largely autonomous reset, (3) learnable but not trivial, and (4) yields a policy that is \textit{robust.} For our dynamic fine grasping problem, we define the following proxy task: \emph{firmly grip a ball swinging in the air}, as Fig~\ref{fig:proxy_task} shows. We attach a ball to a thin fishing line to hang it in the air to allow a swinging motion. The task naturally provides a reset since the string resets the ball around its static position due to gravity, obviating the need for a human supervisor to reset the scene. %However, of interest in this particular task is that (1) grasping this particular object is \emph{representative} of a larger class of problems, and (2) the task is naturally \emph{dynamic}. 

% Finally, the proxy task can be artificial as long as it is representative: one can construct a controller environment in the robot factory, using equipment that would be unavailable during test time, to boost the learning efficiency and robustness \cite{rajeswaran2017learning}.

%We identify the criteria needed to set up a proxy task as follows - (1) (3) learnable, but not trivial. % and (4) exposes learning to diverse and dynamic conditions. 
% \begin{itemize}[noitemsep, topsep=0pt]
%     \item \emph{Reset-friendly}. The usefulness of a proxy task critically depends on how easy it is to reset the environment and to allow RL agents to continue their trial in the real world.
%     \item \emph{Representative}. The strategy and movement for the proxy task should be applicable to various tasks in the broader family of curriculum learning~\cite{graves2017automated,florensa2018automatic,jabri2019unsupervised,rusu2016progressive}.
%     \item \emph{Learnable, but not trivial}. It is possible to trade off different dimensions of challenge when designing the reset task. In our case, we remove the generalization challenge to focus on one single object while raising the precision challenge (picking up a hard-to-grasp small item) and the reactive challenge (exposure to varying dynamics). 
%     \item \emph{Exploiting the freedom at training time}. The proxy task can be artificial as long as it is representative. Using a carefully constructed environment at training time could help simplify reset \cite{rajeswaran2017learning}.
% \end{itemize}

\textbf{Exposing the agent to varying initial conditions enables developing a more robust policy: stochastic reset.} While waiting for a long period to allow the object to come to rest would yield a \emph{static reset}, we show, surprisingly, that resuming sampling with a randomly moving object enables learning of more dynamic and adaptive behaviors. By embracing this type of \emph{stochastic} reset (\texttt{StoR}), we construct a harder-to-learn task, i.e., the agent must react to different initial positions and velocities, so it experiences a larger diversity of conditions during training. As a result, the learned policy becomes more robust to varying dynamic conditions, improving skill transfer to novel disturbances and objects (see Appendix.~\ref{app:system} for quantitative ablation). 

\begin{tcolorbox}[colback=red!5!white,left=1mm,top=1mm,right=1mm,bottom=1mm]
\textbf{Insight:} Designing appropriate proxy tasks to train the agent can simplify the training infrastructure. Additionally, training with stochastic reset can improve the robustness and generalizability of learned policies. 
\end{tcolorbox}

\textbf{Efficient fine-tuning: Improving gradient throughput with asynchronous updates and regularization}

To improve training efficiency in the real world, we care about not only the sample efficiency but also the wall clock time efficiency. In the simulation, operations (e.g., sampling step, resetting hardware) are almost instant. In the real world, however, idle time is unavoidable and adversely affects the speed at which we collect samples: for every second of samples collected, our system spends 4 seconds waiting for a reset. We replace the common training paradigm in simulation, which takes one gradient update after collecting one environment step. We instead perform asynchronous RL updates up to the limit of our computer alongside robot operation. This greatly increases our gradient throughput, effectively committing 10 to 20 updates per data point collected (Update-To-Data, UTD). 

Most off-policy RL methods lower their UTD ratios because additional gradient updates lead to unbalanced training of policy/value functions (i.e., exploding actor / value losses). In this work we find that this problem can be significantly mitigated by using appropriate regularization. By using a standard technique such as layer normalization~\cite{layernorm}, we are able to avoid overfitting and take significantly more gradient steps per data point (See Appendix.~\ref{app:layernorm}). As shown in Fig~\ref{fig:ablation-finetune}, the combination of asynchronous updates and LayerNorm regularization achieve moderately faster training in simulation but, during fine-tuning on the real robot, significantly boosts the efficiency of learning in terms of wall-clock time.

\begin{tcolorbox}[colback=red!5!white,left=1mm,top=1mm,right=1mm,bottom=1mm]
\textbf{Insights.} Improving gradient throughput by leveraging asynchrony and more gradient steps per data point with regularization can make fine-tuning efficient enough for practical real-world use. 
\end{tcolorbox}

\section{Complete System Design for CherryBot}
\label{sec:system}

We combine the preceding design decisions and insights into a single system, CherryBot (Fig~\ref{fig:system}), that can handle challenging dynamic fine  manipulation tasks in the real world. %CherryBot operates in three phases: (1) pre-training in simulation on the proxy task, (2) fine-tuning in the real world on the same proxy task, and (3) deploying on the real world with appropriate state estimation module for downstream tasks. 
Below, we describe the hardware infrastructure and concrete implementation details. 

\textbf{Hardware.}
Fig.~\ref{fig:hardware_overview} shows an overview of our hardware. We built a 6-DOF robot arm equipped with a pair of chopsticks as its end effector, as shown in Fig.~\ref{fig:robot}. Since our hardware is assembled from parts with joints that are not strictly rigid, inaccuracies accumulate along robot links. We document our modeling and system identification procedure in Appendix~\ref{app:hardware}, where we explain how we used a neural network to predict the residual backlash for each joint, achieving position errors $<3$ mm at the robot’s end effector.

\textbf{Reset mechanism.} At the end of every sampled trajectory in the real world (average length is about 80 timesteps or 4 seconds), we conduct a fixed trajectory for reset. We (1) slowly raise the robot arm to a set position, (2) touch the string in the proxy task, expecting to constrain its motion and reduce system entropy, and (3) return the robot arm to a fixed start pose and restart the task. The reset process takes about 16 seconds, during which our policy and value networks keep sampling experiences from the replay buffer and conducting RL updates. Notably, the reset leaves the object in a dynamic condition, introducing the dynamism we need for learning robust policies. 

\textbf{Simulation.} As described in Section~\ref{sec:pretrain}, we leverage simulation for pre-training value functions and policies. We construct the simulation in MuJoCo~\cite{todorov12mujoco} and perform a system identification procedure. We identify the forward kinematics of the arm and associated dynamics parameters for each joint and built a residual neural network to further improve accuracy, as described in Appendix.~\ref{app:hardware}.

\textbf{Perception.} We use different perception modules for training and testing. At training time, we use a motion capture system, Optitrack, to obviate many perception challenges and yield an accurate position for the single object to grasp used in the proxy task (error $\sim0.01$mm). At test time, accurate state estimation may not be available. Our system instead accepts any external perception module that can estimate the object's center of mass, following common practice in visual servoing~\cite{garrett2021integrated,zhang2022visually}. Noticeably, our system does not necessitate a highly precise estimation module w.r.t. the fine manipulation tasks.

Utilizing an external perception module allows our system to be deployed in various downstream tasks. For illustration, we demonstrate our system using (1) a simple segmentation method with smoothing and (2) an off-shelf detection system, YOLO (both documented in App.~\ref{app:system}). When the object is static, easy to recognize and the scene is clean, our in-house module detects the object with little error. However, our scenes can have occlusions and can be changing (e.g., cherry shaking in the wind partially occluded by the leaves), and the perception error could reach $\sim$5mm. In our experiment, we will also inject noise to the perception module's output to demonstrate the robustness of our proposal to perception noise. 

%used a single RGBD camera running at 720p 30hz and  

%After that, we do edge and contour detection to get the center location of the object in pixel space. With the depth measurement at that point, we could project 2D points into 3D using camera intrinsic and extrinsic matrices. The calculated object states will have around 0.3mm offsite at most due to the inaccurate depth measurement and camera calibration error. But, the RL agent could still be robust to this kind of sensor error due to our design choices. [TODO: Check and fix this, shorten it.]
%running at 720p 30hz for tracking the position of the object. In particular, we 

\begin{figure*}[!t]
\centering
    \centering
\includegraphics[width=\linewidth]{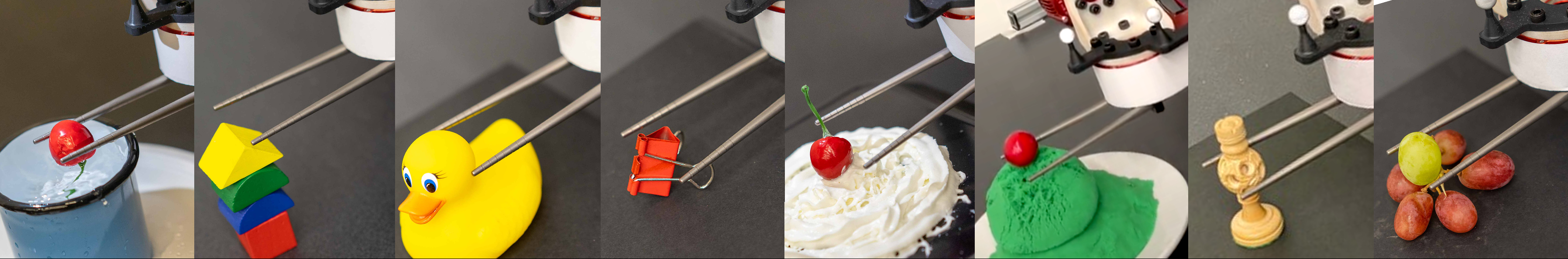}
\caption{\footnotesize{We deployed our system to generalize to a wide variety of objects with different shapes and textures.}}
\label{fig:generalization}
\end{figure*}

\begin{figure}[!h]
\centering
 \begin{subfigure}[t]{.4\linewidth}
    \centering
    \includegraphics[width=\linewidth]{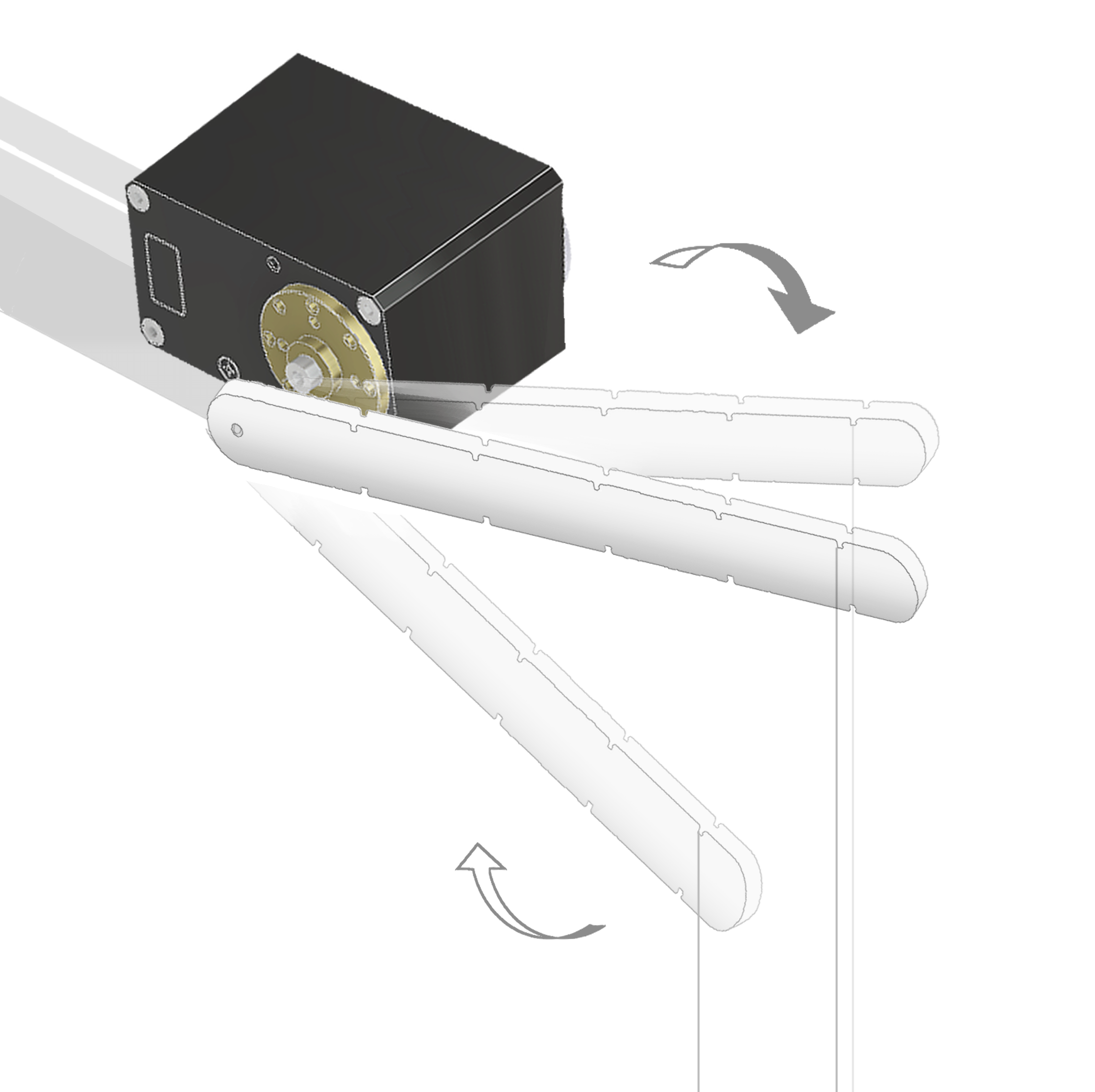}
\caption{\footnotesize{Motor for more dynamic disturbance.}}
\label{fig:motor_disturbance}
  \end{subfigure}
  \hfill
  \begin{subfigure}[t]{.4\linewidth}
    \centering
    \includegraphics[width=\linewidth]{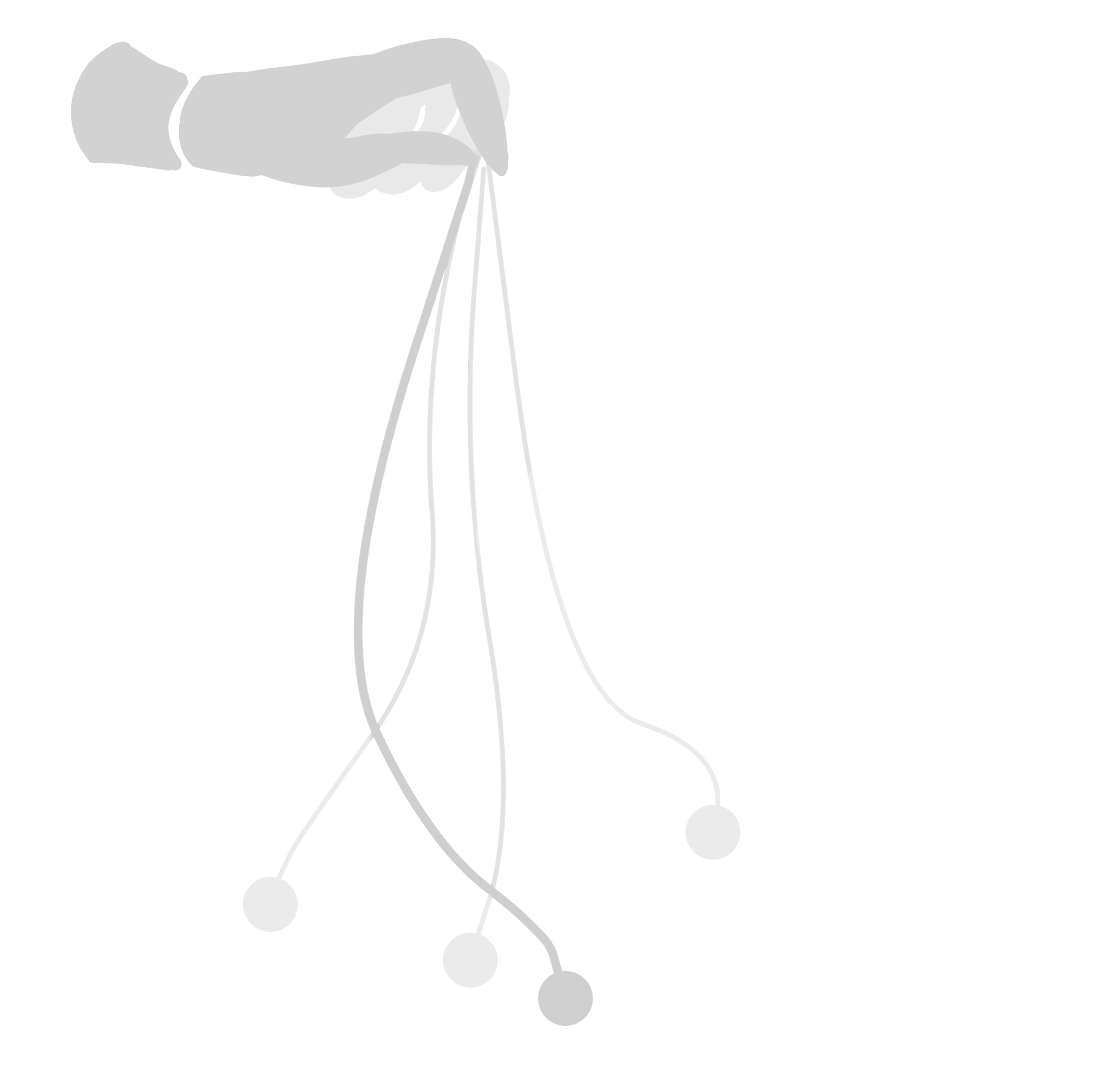}
\caption{\footnotesize{Human disturbance.}}
\label{fig:human_disturbance}
  \end{subfigure}
  \caption{\footnotesize{Dynamic disturbance evaluations we considered. These disturbances test the robustness of the learned CherryBot policies. }}
  \vspace{-1em}
\end{figure}

\textbf{Additional hardware for dynamic disturbance 
evaluation.} 
The goal of our fine grasping policies is to firmly grasp the object, despite dynamic disturbances. It is challenging to test this type of robustness without additional machinery to introduce disturbances. To this end, we design a motor disturbance mechanism to systematically inject dynamic disturbance so we could evaluate the robustness of agents. As shown in Fig~\ref{fig:motor_disturbance}, the motor perturbs the motion of the ball by rotating the motor's arm and dragging the string attached to it. By varying the string's hanging point, the object can be pulled with different sizes of disturbance (from smaller disturbances to bigger ones, denoted as $20 \sim 100$).

\section{Evaluation}

\begin{table*}[t!]
    \normalsize
    \setlength{\tabcolsep}{3pt} % Default value: 6pt
    \centering
    \resizebox{\textwidth}{!}
    {%
    \begin{tabular}{rrrr|rrr|rrr|rrr|rrr}
    \toprule
             &\multicolumn{3}{c}{\textbf{Marble ball}}& \multicolumn{3}{c}{\textbf{Cherry}}
             &\multicolumn{3}{c}{\textbf{Gaussian noise}} &\multicolumn{3}{c}{\textbf{Static ball}} &\multicolumn{3}{c}{\textbf{Dynamic ball}}

             \\
    \midrule
    & Rew. & Succ. & TTS & Rew. & Succ. & TTS & Rew. & Succ. & TTS & Rew.
    & Succ. & TTS & Rew. & Succ. & TTS
    \\
    \midrule
    \rowcolor{Gray}
    Our RL  & \textcolor{blue}{\textbf{242.3}} &
    \textcolor{blue}{\textbf{100\%}} & \textcolor{blue}{\textbf{1.77}} & \textcolor{blue}{\textbf{321.7}} &
    \textcolor{blue}{\textbf{100\%}} & \textcolor{blue}{\textbf{0.92}}  & \textcolor{blue}{\textbf{130.4}} &
    \textcolor{blue}{\textbf{90\%}} & \textcolor{blue}{\textbf{2.06}}& \textcolor{blue}{\textbf{565.3}} & \textcolor{blue}{\textbf{100\%}} & \textcolor{blue}{\textbf{0.28}} & \textcolor{blue}{\textbf{512.3}} &
    \textcolor{blue}{\textbf{100\%}} & \textcolor{blue}{\textbf{0.48}}\\
    20HZ VS & 64.3 & 60\% & 1.12 & 113.1 & 70\% & 1.17 & -30.9 & 50\% & 2.29 & 196.2 & 100\% & 1.49 & 183.5 & 90\% & 1.38 \\
    100HZ VS  & 48.4 & 50\% & 0.93 & 110.7 & 60\% & 0.32 & 109.3 & 60\% & 0.64 & 535.0 & 100\% & 0.49 & 255.0 & 90\% & 0.49 \\

    Human &-&-&-&-&-&-&-&-&-& -38.8 & 100\% & 2.20 & -100.2 & 100\% & 2.86\\
    Replay &-&-&-&-&-&-&-&-&-& -56.9 & 80\% & 1.45 & -218.4 & 50\% & 2.29 \\
    BC &-&-&-&-&-&-&-&-&-& -354.4 & 30\% & 0.58 & -487.0 & 10\% & 0.48 \\
    \bottomrule
    \end{tabular}
    }
    \caption{\footnotesize{Evaluation results, including average reward (Rew.), success rate (Succ.), and time-to-success for only \textbf{successful} trials (TTS in seconds). Note the RL agent is able to adapt to its mistakes and would retry after failure rapidly, therefore achieving a high success rate within the allocated 6 seconds.}}
    % \kay{What is our insight? (1)identify interesting numbers(2) provide intuition (3) refer to video on web. I think worth talking about (1) Static ball, we are much faster, but similar reward. is it because we jump faster than 100CT <> dynamic ball, we are a tiny bit faster but much higher rew. (2) we on average spend longer picking cherry/marble/noise but that is because we retry for trials that CT runs out of time and failed. (3) 20Hz CT is pretty damn good look at the graph.. } }
    \label{tab:static}
\end{table*}

In our experiments, we evaluate our proposal on fine manipulation challenges (grasping slippery balls or various small items) and during dynamic disturbances (created using programmed motor movements or human interference). We intend to answer the following questions: (1) Can a policy learned by CherryBot be robust and reactive to dynamic scenarios? (2) Can our policy generalize to different objects? (3) Is our proposal robust across random seeds? (4) How do our design decisions impact the final results for CherryBot?
% (3) Are we cherry-picking on our cherry-picking agent? 

\subsection{Tasks}
\label{eval:task}
We test the \emph{reactivity}, \emph{robustness}, and \emph{generalization} of our agent quantitatively and qualitatively. View the video recordings on our website for visualization of our evaluations.

\textbf{Reactivity to dynamics in the scene.}
In natural scenarios, a small object might be moved by external disturbance, e.g., a leaf flying in the wind. We simulate this kind of disturbance by installing a motor that pulls the string that suspends the object (Fig.~\ref{fig:motor_disturbance}). Additionally, we introduce a more challenging, more spontaneous source of external interference---humans. As shown in Fig.~\ref{fig:human_disturbance}, a person would drag the string while the agent tries to grasp and shake the ball attached to it. We provide a qualitative video on the website reporting the performance of our agent on the latter task. %\footnote{Due to the random nature of the human disturbance, it is hard to give a ``fair'' comparison across agents. Hence, we only report our RL agent performance for this task. Please check our website for videos.}

\textbf{Robust to perception noise.}
In the real world, perception errors are inevitable due to,  e.g., lightning conditions or occlusions. We simulate such errors by injecting noise into the tracked positions of the object, i.e., adding a small Gaussian noise to the output of our Optitrack system and using this noisy position as the states~\cite{underactuated}.

\textbf{Generalization. }
Different objects have varying shapes, sizes, and textures. We first conduct a quantitative evaluation of our system's generalization ability on grasping hanging cherries with varying shapes and sizes and a hanging marble ball with a slippery surface that made grasping with chopsticks taxing even for humans~\cite{ke2020telemanipulation}. Further, we conduct qualitative evaluations on an assortment of objects and dynamic scenarios (Fig.~\ref{fig:generalization}), grasping objects of varying shapes (clip, chess), sizes (duck, puzzles) and textures (chess, grapes) and with varying non-rigid support (wind disturbance, floating water, melting cream, falling sand). We further included experiments using an off-the-shelf detection system, YOLO, which empowered our system to grasp objects with more diverse shapes and colors, shown in App.~\ref{app:yolo}.

\textbf{Ablation.} During training, we employ a high-accuracy tracking cage that eliminates perception errors, making it an ideal platform for ablation testing. We first place the tracker ball at a fixed position and ask the agent to grasp the \textbf{Static} ball, examining how precisely the agent can grasp. Second, we hang the ball and give it a random initial velocity (\textbf{Dynamic} ball), testing the agent's reactivity to dynamic scenes.

% visualizes a series of objects we evaluate our system on. The objects vary in shapes, sizes, textures. Some of the properties (e.g. the slippery surface of the marbal ball) can make the .  %In the real world, the challenges of reactivity, robustness, and generalization are often entangled together. Thus we try to test the agent's performance on these various objects.
%To evaluate our agent's generalization ability across even more objects, w

%Further, we also test the agent with qualitative tasks to show the agent's generalization ability across different surface support, object shapes, and disturbance. As shown in fig[], from left to right, we shown that our system is able to 1) grasp a cherry from a tree branch swaying in the wind, pick up a cherry from the top of floating sand, pick up tomatoes from a cluster of tiny grapes 2) pick up objects with various shapes (cheese, clamp, toy duck, etc), and 3) pick up a moving ball under human disturbance.

\subsection{Baselines}

We evaluate our agent on five candidate baselines: (1) 20HZ VS, our heuristic-based controller that we used to collect demonstrations, and (2) 100HZ VS: our heuristic controller with a different set of gains. Optionally, we test (3) Humans: a human expert teleoperation the robot, (4) Replay: replaying successful human demonstrations, and (5) Behavior Cloning (BC): a practical imitation learning algorithm~\cite{pomerleau1988alvinn} trained with our collected data, detailed in App.~\ref{app:il}. 

For each task and each agent we selected, we test over 10 trials and summarize performance statistics. In total, we conducted more than 500 trials. For fair evaluation and to accommodate the slow motion of the human baseline, we use a fixed horizon (6 seconds) at test time.%\footnote{The difference in time horizon explains why the rewards during training and testing are slightly different}. 
We determine the success of the trajectory by inspecting whether the robot could securely lift the object and stabilize it at a height of 0.1m in the world frame. 

\subsection{Reactivity to dynamic disturbance}

\begin{figure}[htbp]
  \centering
\includegraphics[width=0.7\linewidth]{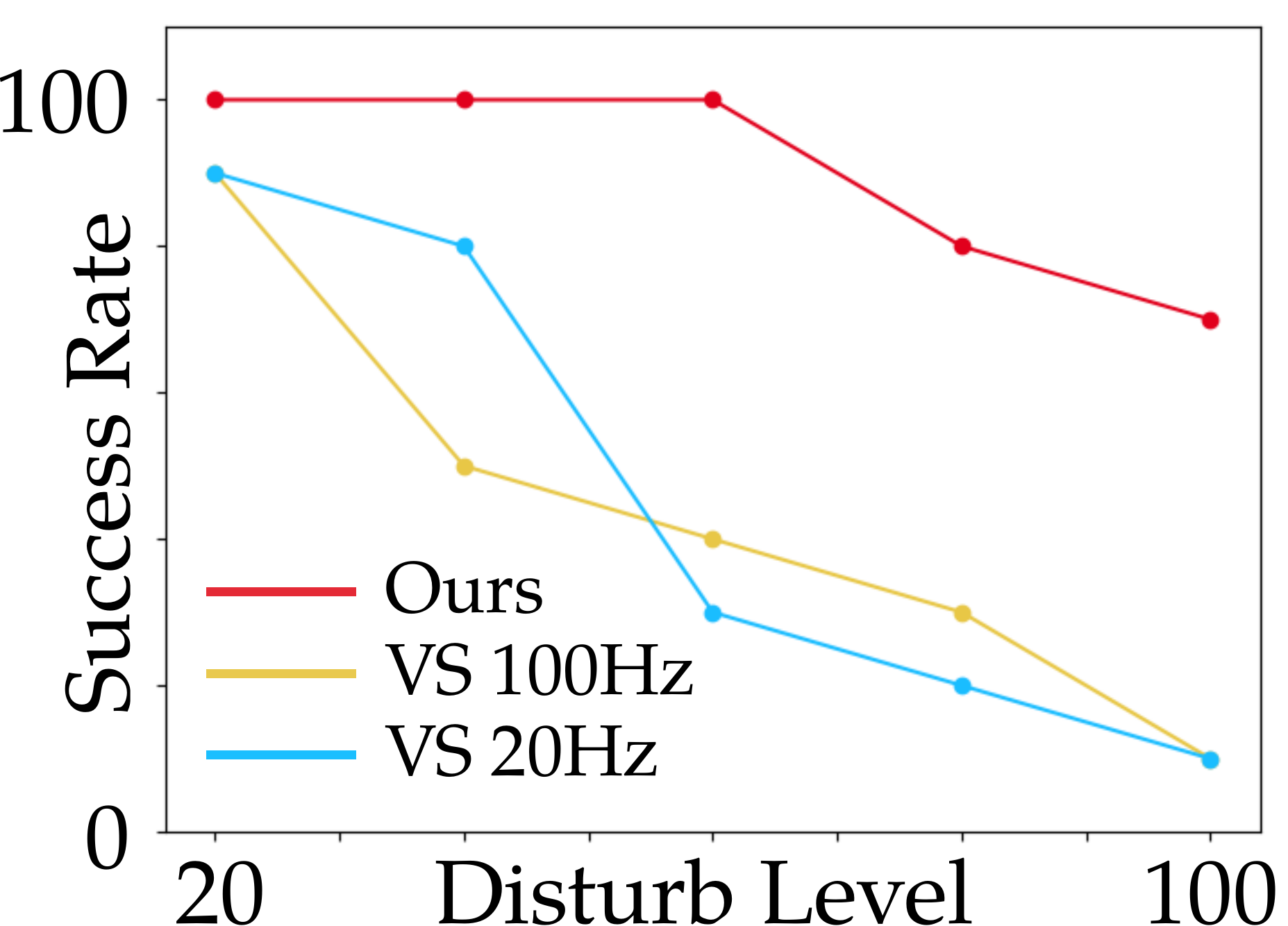}
  \caption{\footnotesize{Success rate under different disturbance. The CherryBot controller's RL agent is robust to disturbances that significantly exceed those for baseline methods.}}
  \label{fig:result-reactivity}
  \vspace{-0.5em}
\end{figure}

Fig \ref{fig:result-reactivity} shows the agents' performance under varying degrees of dynamic disturbance injected by the motor dragging. We observe that our RL agent outperforms the baselines by a large margin across all degrees of motor dragging disturbance. In the human disturbance task, our RL agent succeeded in 17 of 29 trials of human disturbance within the time limit. % We note that the  could be moving up to X meter / s by human dragging. %Compared to VS baselines, which are sensitive to dynamic disturbance and could be reactive only with scant external perturbations, our agent trained on \texttt{Proxy} with hybrid controllers is more reactive.

\subsection{Robustness to noise}
As shown in Table \ref{tab:static}, when Gaussian noise is applied to the sensor, the RL agent performed the best,  achieving a 90$\%$ success rate, 
%\sandy{Below, state which VS achieved which rate.}
significantly outperforming the baseline VSs (20Hz achieved 50$\%$ and 100Hz achieved 60$\%$ success rates). The noisy state estimation is fatal to our vanilla VS baselines which took hours of gain tuning, and it will require further tuning or adding of adaptive modules to improve performance. In contrast, our agent was naturally robust to those noises due to our design choice of \texttt{SimR}, \texttt{Proxy}, and \texttt{StoRe} to train the agent on more challenging tasks before deployment.

\subsection{Generalization}
Table \ref{tab:static} displays our quantitative results on hanging marble balls and cherries, measuring the rewards, success rate and time-to-success on picking up these objects. Our agent significantly outperforms the baselines; the baselines succeeded in some trials quickly, but they could not readjust to their failures to complete the task within the allocated time and instead kept batting the object around. 
\begin{table}[!h]
    \centering
    \resizebox{\linewidth}{!}
    {
    \begin{tabular}{cccccc}
    \toprule
    \textbf{Task} & \textbf{Succ. (\%)} &
    \textbf{Task} & \textbf{Succ. (\%)} &
    \textbf{Task} & \textbf{Succ. (\%)}
        \\
        \midrule
        Water Cherry  & 60 (9/15) & Puzzle  & 100 (10/10) & Duck  & 50 (5/10) 
         \\
        \midrule
         Clip  & 30 (3/10) & Cream Cherry & 100 (10/10) & Sand Cherry  & 53.3 (8/15) 
        \\
        \midrule
        Chess  & 80 (8/10)  & Grapes  & 70 (7/10) & Tree Cherry & 40 (4/10)
        \\\bottomrule
    \end{tabular}
    }
    \captionof{table}{\footnotesize{Success rates of generalization tasks. CherryBot learned policies and was able to generalize non-trivially across objects.}}
    \label{tab:score-real}
    \vspace{-1em}
\end{table}

We further evaluate our agent on more diverse generalization tasks in Fig~\ref{fig:generalization} and show the success rate in Table~\ref{tab:score-real}. Our   
 agent's overall success rate was $64\%$ (of 100 trials), even with our simple perception module. We summarize in four categories the factors explaining the low success rates on some generalization tasks: (1) the object has no firm contact point and keeps slipping out (clip and chess), (2) the object size is too big (duck), (3) the failure of grasping is fatal (cherry drops in the falling-sand task), and (4) perception errors due to occlusion (tree cherry-pick). %By embracing interchangeable \texttt{Vis}, our agent shows great generalization potential. %\yc{kay,shall we add more factors here? and change the first table a bit}

\subsection{Further analysis}
We examine the performance of all baselines on our proxy task, which removes perception noises but tests the agent's precision and dynamic reactivity. Table \ref{tab:static} presents results on proxy tasks in both static and dynamic modes. In general, our RL agent outperforms all baselines in success rate, time to succeed, and total rewards. Both the 20Hz and 100Hz VS baselines could also achieve 90 to 100$\%$ success rates, respectively, on the proxy tasks, albeit taking a longer time to succeed than our RL agent. It is worth noting that our RL agent, which runs at a hybrid control frequency, uses less time to succeed compared to 100HZ VS. This implies that our hierarchical framework with its low-frequency learned policy develops a more efficient strategy for grasping in the dynamic scene than our hand-designed controller. 

\textbf{Are we cherry-picking a cherry-picking agent?} To clarify the reproducibility of our proposal: we use an open-sourced codebase d3rlpy~\cite{d3rlpy}, with default implementations and hyperparameters, to run our ablation studies in the simulation and our real-world robotic experiments. We conducted 5 random seed sweeping in simulator ablation studies and 3 random seed sweeping in real-world fine-tuning studies, verifying the the efficacy and reproducibility of our proposal. See Appendix~\ref{app:sweep} for details.

 % 3. asynchronous training is good

% Qualitative comparing: disturbance.

\section{Conclusion}
We present a system, CherryBot, for learning robust dynamic fine grasping in unstable conditions. We provide empirical evidence demonstrating the effectiveness of our proposed system on a low-cost chopsticks-equipped robot and validate its efficiency, reactivity, robustness and potential to generalize, showing how reinforcement learning can be a practical and competitive tool to learn dynamic and precise behavior.

%We present a system for learning dynamic fine manipulation with chopsticks in the real world. We show how reinforcement learning can be a useful tool to learn dynamic and precise behavior. We show how a combination of careful design decisions for pretraining in simulation, combined with finetuning on the real world with a dynamic proxy task can lead to robust and reactive learned behavior. We show the efficacy of this system on a low-cost real-world chopsticks robot, and empirically validate different design decisions in the same. 

%Though we were able to successful accomplish a handful of grasping tasks in the real world, there are still numerous challenges that would make for exciting future work. For instance, going from the tracking system that we are using now to a full vision based RL system would increase applicability. Additionally, extending our work to a broader diversity of tasks, including long horizon tasks would open up a wider range of tasks and data. Developing a deeper theoretical of why these design decisions proved so important, would also be very interesting.

Despite successfully performing a diverse set of grasping tasks in the real world, there remain significant challenges to address in future work. Expanding our system to a wider range of tasks, including those with longer time horizons, could broaden its applicability. Upgrading the current perception module to include object pose and to have an end-to-end vision-based RL system could enhance its usability and allow further fine-tuning after being deployed. Exploring the theoretical reasons behind the importance of our design decisions, or providing insights into their applicability to different hardware or domains, would also open fascinating areas of future study.

% such as the horizon restriction for RL that could be applied and the sample efficiency problem. Also, perception and actuator noise  

% Challenge of horizon \& sample efficiency for manipulation versus locomotion

% locomotion = 35Hz, continuous smooth reward

% ours = 20Hz, discrete reward, long horizon issue.
\section*{Acknowledgment}

We thank
Kendal Lowrey and Collin Summers for helping with system identification, Emo Todorov for gifting access to Mujoco, Chris Atkeson, Vikash Kumar for feedback on the draft, 
Mira Chew and Selest Nashef for 3D design and printing.

This work was (partially) funded by the National Science Foundation NRI (\#2132848) and CHS (\#2007011), DARPA RACER (\#HR0011-21-C-0171), the Office of Naval Research (\#N00014-17-1-2617-P00004 and \#2022-016-01 UW), and Amazon.

%% Use plainnat to work nicely with natbib. 

% \bibliographystyle{unsrtnat}
% \bibliography{main.bib}

%\clearpage

% -----------------------------------------
\section{Appendix}
%\sandy{Note that I did not do a second review of the Appendix.}
\subsection{Hardware}
\label{app:hardware}

We document how we built a simulation via kinematic modeling and a system identification procedure. We explain how we used a neural network to predict the residual backlash to achieve position errors $<3$ mm at the robot’s end effector. We also elaborate on our perception module at test time.

\paragraph{Modelling and Kinematic Calibration}

Performing fine-motor skills requires a carefully calibrated kinematics model that reflects the hardware setup in the real world. The robot manufacturer provides a kinematics model, but this (a) does not account for the chopsticks end-effector, and (b) is not tuned to the degree of accuracy that we require. Therefore, we employ data-driven calibration pipelines that allow us to achieve highly accurate position estimates from joint angles.

The Optitrack cage provides precise and accurate pose information, which we can leverage as ground truth data. When we mount a pair of chopsticks to the robot end-effector, one of them is fixed (``primary'' chopsticks) and the other one can be rotated by the end effector joint (``moving'' chopsticks). By placing a tracker on the tip of the primary chopstick, we can measure the end-effector position in the 3D space. 

First, we need to measure the robot's base position and orientation on the table. We spin the base joint of the robot for several rotations in both directions. This causes the tracker to trace out a circle several times. By measuring the position and inclination of the circle, we can determine the tilt and position of the robot on the table.

We then use teleoperation to control the robot to move inside the workspace, recording the end-effector position and the robot joint angles. We then use this data with either a black-box or gradient-based optimizer (using the factory defaults as the initial guess) to solve for a set of accurate parameters for the kinematics model by minimizing the FK loss. To regularize the optimization, we find that it is important to penalize deviating too far from the manufacturer-given kinematics model. We observed the kinematic error being $1.2 \sim 3$ mm at the end effector tip in the task space we tested.

% @abhay. FK IK.  Feel free to copy Github readme etc

\paragraph{Residual Estimation to Improve Kinematics}

To combat inaccuracies in our kinematics, particularly due to backlash in the arm joints, we train a model to predict the backlash in each joint of the arm as a function of the current joint angles. The intuition behind this is that in certain orientations the backlash of the joints may induce errors that are predictable. These dynamics wouldn't be able to be captured by the DH parameters that parameterize the arm kinematics, but a nonlinear function approximator like a neural network would be able to predict this.

The residual estimator network is trained on a dataset of joint angles and grounds truth positions (as determined by the mocap cage) collected in and around the workspace. Additionally, the predicted backlash is constrained to be at most 10\% more than the figure listed by the manufacturer, to prevent deviating too far from the kinematic model. In practice, we found that this approach reduced the average position error from ~1.8mm to ~1.5mm, about a 17\% reduction, which is significant given the high-precision nature of the task.

\paragraph{System Identification}
\label{app:sysid}

To build a simulator with dynamics as close to real as possible, we also employ black box optimizer to fit the dynamic transition in our simulator to the real world using recorded trajectories of the robot in and around the workspace. we optimize the environment's parameters (centers of mass of joint bodies, friction coefficients, contact solver parameters, etc.) in order to minimize the divergence of simulated rollouts from the ground truth recording. Again, we regularize this optimization by penalizing the divergence of the simulator parameters from the initial guess.

To run the optimization program efficiently, the code leverages the fast computational speed of Julia and the Lyceum MuJoCo wrapper~\cite{summers2020lyceum}. We do not expect that this optimization pipeline will result in a simulator that matches reality exactly because of the limitations of the simulator: for example, the shape of the chopsticks and contact parameters are simplified. But it does yield good enough parameters to enable simulator pre-training.

% @abhay + kay

\paragraph{Perception Module}

Our system can accept any perception module at test time as long as it yields an estimate of the center of the mass of the object to grasp. For demonstration, we built a heuristic-based perception module that estimates the object's centroid position from an Azure Kinect camera.
It uses salient pixels to estimate the object’s CoM position from RGB-D streams. At test time, the user chooses an object to grasp and specifies a salient color (e.g., red for cherries). We first filter out noise in the image using a Gaussian blur and mask the salient region similar to the desired color. With edge and contour detection, we can get the centroid of the object in pixel space. We then project the 2D point location to 3D using camera intrinsic and calibrated extrinsic matrices.

When the object is static, easy to recognize and the scene is clean, our current perception module detects the object with little error - detecting a red marble on a white background has a ~0.3mm error. However, in our experiments, the scenes can have occlusions and can be changing (e.g., cherry shaking in the wind partially occluded by the leaves), and the perception error could reach $\geq$ 5 mm. 

We conduct ablation experiments to verify the robustness of our system to perception noise: After injecting a Gaussian noise with a mean of 5 mm into the perception module, the performance of the visual servo controller dropped from 90\% to 50~60\%, shown in Table II. Our system. however, achieved 90\% even with these noises present, highlighting its robustness to perception noise.

\paragraph{Policy Input and Output} Our agent follows~\cite{ke2021grasping} and uses the end effector pose (i.e., 3D vector of x-y-z position and 4D quaternion of rotation) plus the perception info (i.e., 3D vector of the center of mass) as input. The high-level policy (20Hz) outputs a command effector pose (3D xyz + 4D rotation), which will be translated by an Inverse Kinematic solver to be a 7D vector of commanded joint positions for the whole arm, including the end effector. Each positional joint command is fed to a lower-level PID controller (1000Hz).

\subsection{System Design Ablations}
\label{app:system}

All ablation experiments mentioned in \ref{sec:method} were run with the same experimental design to make fair comparisons. Unless otherwise specified in the paper, we focus on the SAC implementation provided by the d3rlpy library, using the default hyperparameters in d3rlpy library, sweeping across the seeds 120, 121, 122, 123, and 124. When run in simulation, we use the simulator whose tuning process is described in Sec. \ref{app:sysid}, and real-life experiments are run with the same hardware and perception as described in Sec. \ref{sec:system}.

Aside from the ablations studied in depth in Section \ref{sec:method}, we also performed a few more experiments validating other design choices.

\paragraph{Update-to-Data Ratio}

\begin{figure}[htbp]
\centering
\includegraphics[width=0.8\linewidth]{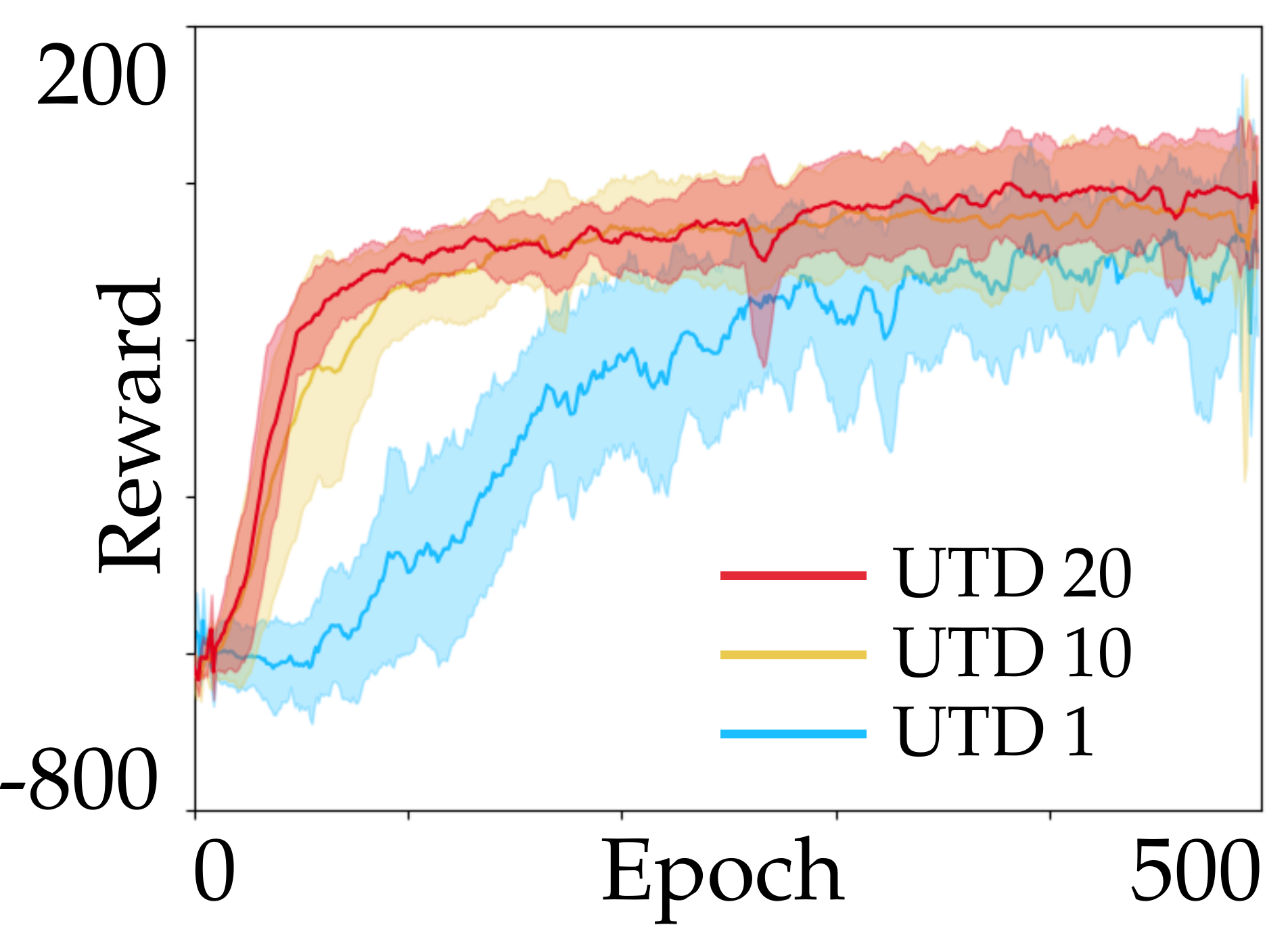}
    \caption{Training our RL agent in the simulator with varying values of UTD. Higher values of UTD converge much faster than lower ones.}
    \label{fig:ablation-utd}
\vspace{-1.5em}
\end{figure}

The UTD (Update-to-Data ratio) controls the ratio of gradient steps per environment step. Our system achieves an effective UTD=10 through asynchronous updates. We illustrate in Fig. \ref{fig:ablation-utd} that higher UTD values of 10 and 20 result in much faster convergence than the traditional choice of UTD=1. Running more updates for the same amount of data seems to drastically increase the sample efficiency and expedite learning. The key takeaway is that a large UTD ratio can be favored for its practicality in real-world training with real-time constraints and it showed empirical improvement in learning speed, making real-world learning more tractable.

\paragraph{LayerNorm}: \label{app:layernorm} LayerNorm is a simple regularization that turns out to be very effective when we employ a high UTD ratio. To show the impact of Layer Norm regularizer, we run trials for it with different UTDs.

\begin{figure}[htbp]
\centering
  \begin{subfigure}{.24\textwidth}
    \centering
\includegraphics[width=\linewidth]{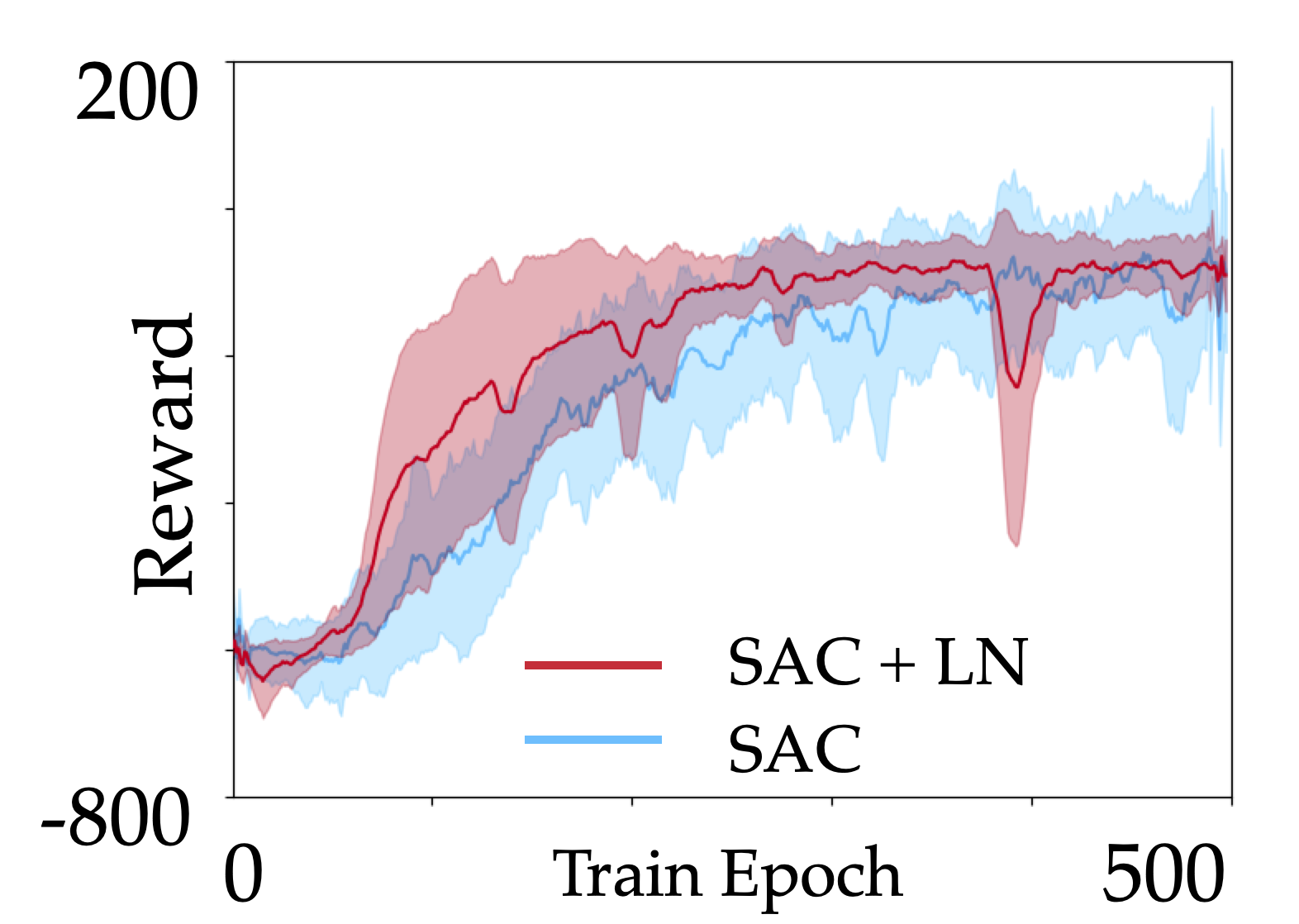}
    \caption{\footnotesize{UTD=1}}
  \end{subfigure}
  \begin{subfigure}{.24\textwidth}
    \centering
\includegraphics[width=\linewidth]{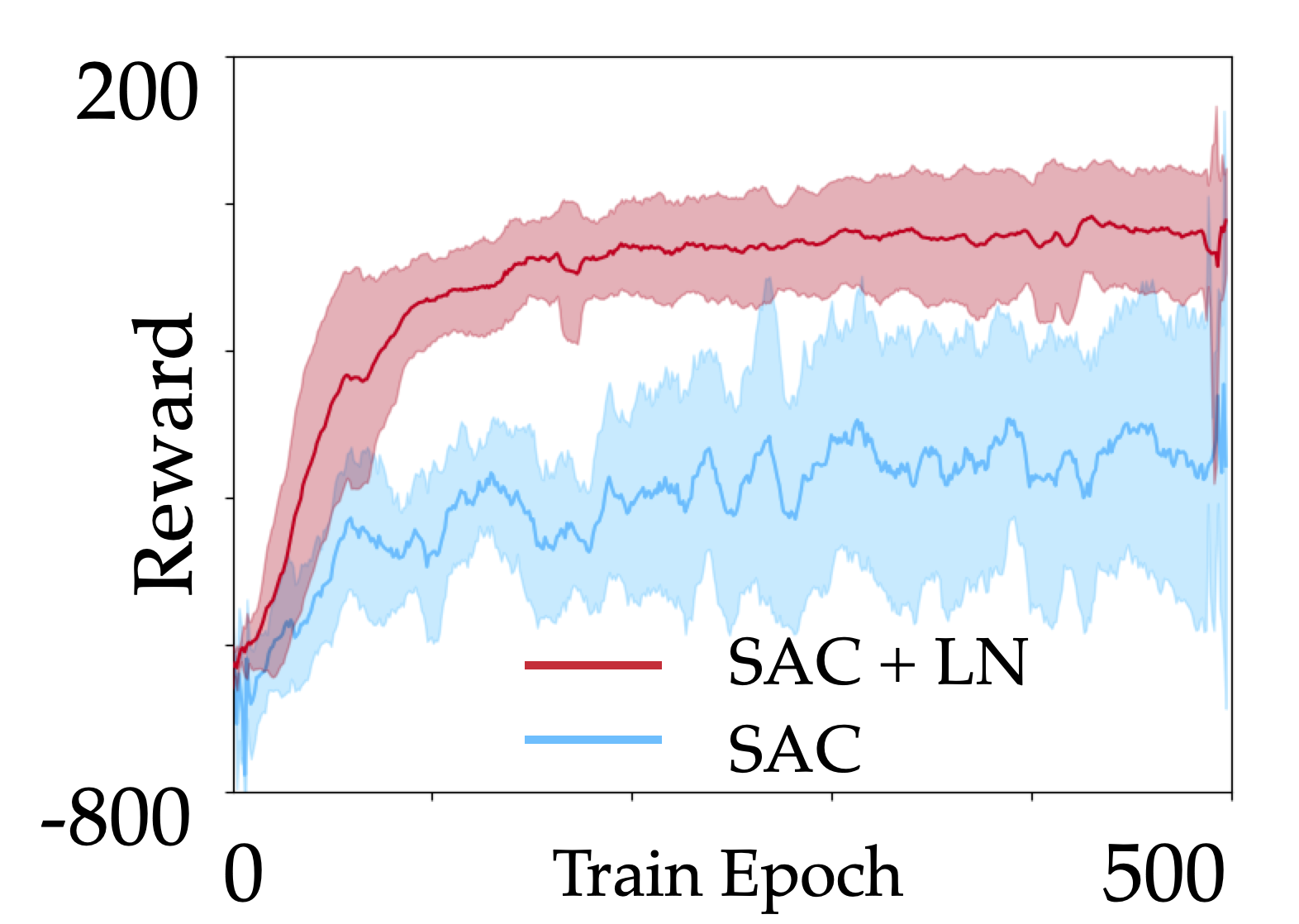}
    \caption{\footnotesize{UTD=10}}
  \end{subfigure}
\caption{\footnotesize{Analysis of using LayerNorm regularization at different UTD}}
\label{fig:layernorm}
\vspace{-1em}
\end{figure}

\paragraph{Stochastic and static reset performance}

Our proxy task features a ball swinging from a string which, notably, has a stochastic reset, as described in Section \ref{sec:finetune}. We illustrate the impact of this design choice via an ablation study in the simulator, training two agents with static or stochastic reset for the same number of epochs. During the evaluation, we test the two agent's performance on the static-reset task and stochastic-reset task. The one trained with stochastic reset yielded an average score of \textbf{198.8} on 10 trials, significantly higher than the agent trained with static reset, which yielded only \textbf{127.8}. Evidently, the use of stochastic reset, which yields a larger and more varied initial state distribution, promotes a better-performing policy.

% @abhay overall describe the system ablation study design. Why we do this. testing verifying design choice. We run things in the above sim. We consider SAC IQL from d3rlpy. Default hyperparameters. 5 random seeds.

\subsection{Task Design}
\label{app:task}

We use a proxy task (as described in the main paper) as a representative task during our agent's fine-tuning stage and then solve other generalization tasks in zero-shot at deployment time. Instead of ``catching" the moving objects directly, we move the agent to be close to the target first and then manipulate it. Thus, the challenging task was divided into two stages: 1) the simple, slow-reaching stage and 2) the reactive grasping stage. During the first stage, assuming the position of the moving objects is known, we use a heuristic policy to approach the object slowly to be within about $\sim 3$ centimeters away (as the object moves, the distance is not guaranteed to be upper bounded). Then, we will use the learned RL policy or our baseline policies to solve the task in the reactive grasping stage.

\textbf{Success criteria}: The task is considered successful if the robot is able to firmly grasp and lift the correct target object. In the real world, we also used the force torque sensor on the robot joint (which yielded noisy readings) to judge the success of the grasp.

\textbf{Reward design}
Prior RL works had reported using both sparse and dense rewards for manipulation tasks to encourage a desired motion \cite{rajeswaran2017learning,riedmiller2018learning,zhu2019dexterous}. Our reward function also consists of two parts: dense rewards and sparse rewards. The various dense and sparse terms of the reward function are described in Table \ref{tab:rewards}.

\begin{table}[h!]
    \normalsize
    \centering
    \begin{tabular}{rl}
    \toprule
    \multicolumn{2}{c}{\textbf{Dense Rewards}} \\
    \midrule
    Weight & Description \\ 
    \midrule
    -5 & $\lVert \cdot \rVert_2$ between object and the chopsticks\\
    -10 & $\lVert \cdot \rVert_2$ between object and goal\\
    +1 & Correctness of distance between chopstick tips \\
    +2 & Squeezing torque of chopstick \\
    \midrule
    \multicolumn{2}{c}{\textbf{Sparse Rewards}} \\
    \midrule
    -5 & Object stayed around initial position \\
    +5 & Firmly grasp the object and lift to goal\\
    -10 & $\lVert \cdot \rVert_2$ between object and chopsticks $>$10cm\\
    \bottomrule
    \end{tabular}
    \caption{Rewards design}
    \label{tab:rewards}
\end{table}

\subsection{Heuristic controller design}
\label{app:heuristic}
We design our VS(visual servo) controller following human's grasping philosophy in a state-machine manner. Similar to our system, the VS controller also relies on an external state estimation module. First, the robot will align the centroid of the object with the center of chopsticks tips using PD control in 3 axes. It adjusts its pose until the position offset is within a small threshold (1mm), and will then close the chopsticks to attempt the grasp. After that, it will move the object to the goal position with a proportional positional controller. If the robot cannot maintain the height offset within 1 mm during any state, we will return the policy back to the alignment state. The heuristic-based VS controllers are sensitive to gain-tuning and precision of the perception. We spent a reasonable amount of hours tuning the gains of our controller, but there perhaps still remains room for improvement.

\subsection{Imitation Learning}
\label{app:il}
In our early exploration, we experimented with human teleoperators and simple imitation learning. We noted that the replay of the successful demonstration did not have a 100\% success rate, suggesting that fine manipulation requires a precise and reactive policy to be robust to potential sensor and actuator errors. 

We follow~\cite{ke2021grasping} for imitation learning baseline. For the replay, we sample 20 out of 100 successful trajectories generated by human teleoperation. For behavior cloning, we only used the 100 successful trajectories in training. We follow the convention of imitation learning \cite{ke2021grasping, peng2020learning} and filter out failed trajectories.

Surprisingly, in both static and dynamic ball setups, BC's performance is even worse than replay, suggesting that the performance of imitation learning is heavily dependent on data support. It is hard to apply BC on these fine manipulation tasks because a tiny divergence between train and test will shift the agent far from a success, motivating us to seed a self-supervised learning method that can learn from trial and error.

\subsection{Off-shelf Perception Module}
\label{app:yolo}
Our system can plug in any external separate perception module that returns the center-of-mass (CoM) of the object and does not necessitate a high-accuracy perception module. To showcase the robustness of our system, we include experiments using an off-the-shelf detection system, YOLO~\cite{bochkovskiy2020yolov4}, on our website. YOLO allows our system to grasp objects with more diverse shapes and colors. To use YOLO, we calculate the center of the bounding box returned by YOLO to feed to our system. We ask our system to pick up some plastic figures, exemplifying how to apply our system to handle objects with more complicated shapes and shades. 

%We also tried Segment Anything SAM as an alternative to extract the Center of Mass. However, our compute resource (a single Nvidia 3070 GPU) limit the speed at which we can run the perception module in close-loop. We only show the open-loop segmentation result below. We recognize that detecting CoM or object pose can be an open research question and that our system’s performance would be influenced by the quality of the CoM estimation. However, employing a separate state estimation system allows us to leverage advances made by other researchers. The success of these foundational segmentation models demonstrates the practicality of our assumption that estimating CoM for arbitrary objects can be feasible, which allows our system to benefit from advances in object detection and pose estimation research.

\subsection{Generalization tasks and their emphasis}
\begin{table*}[!h]
    \centering
    \resizebox{\linewidth}{!}
    {
    \begin{tabular}{cccccc}
    \toprule
     \textbf{Figure} & \textbf{Task} & \textbf{Succ. (\%)} &\textbf{Figure} & \textbf{Task} & \textbf{Succ. (\%)}
        \\
        \midrule
        \begin{minipage}[b]{0.3\columnwidth}
		\centering
		\raisebox{-.5\height}{\includegraphics[width=\linewidth]{media/focus-water.png}}
	\end{minipage} & Water Cherry & 60 (9/15) & 
              \begin{minipage}[b]{0.3\columnwidth}
		\centering
		\raisebox{-.5\height}{\includegraphics[width=\linewidth]{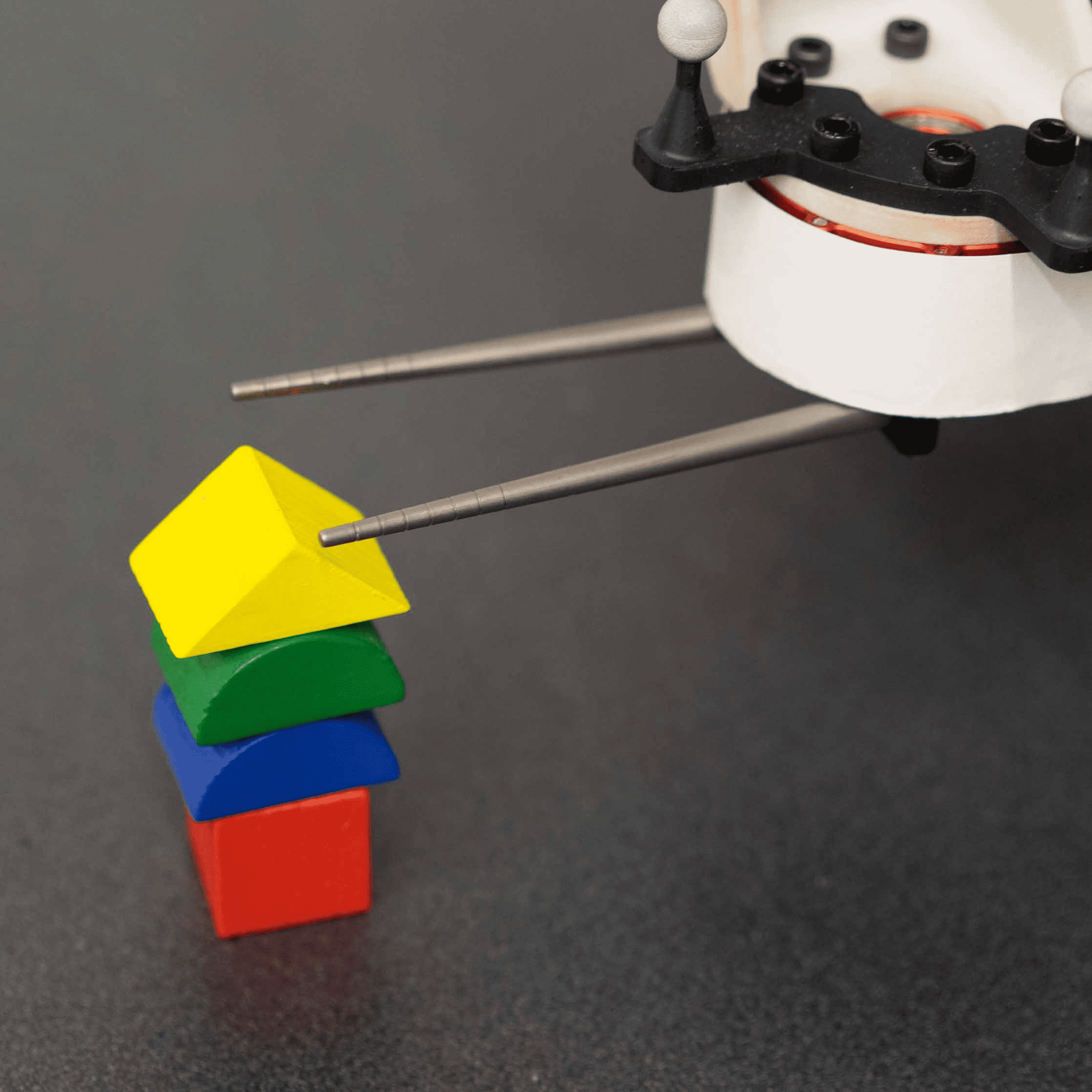}}
	\end{minipage} & Puzzle  & 100 (10/10)  \\

        \midrule
          \begin{minipage}[b]{0.3\columnwidth}
		\centering
		\raisebox{-.5\height}{\includegraphics[width=\linewidth]{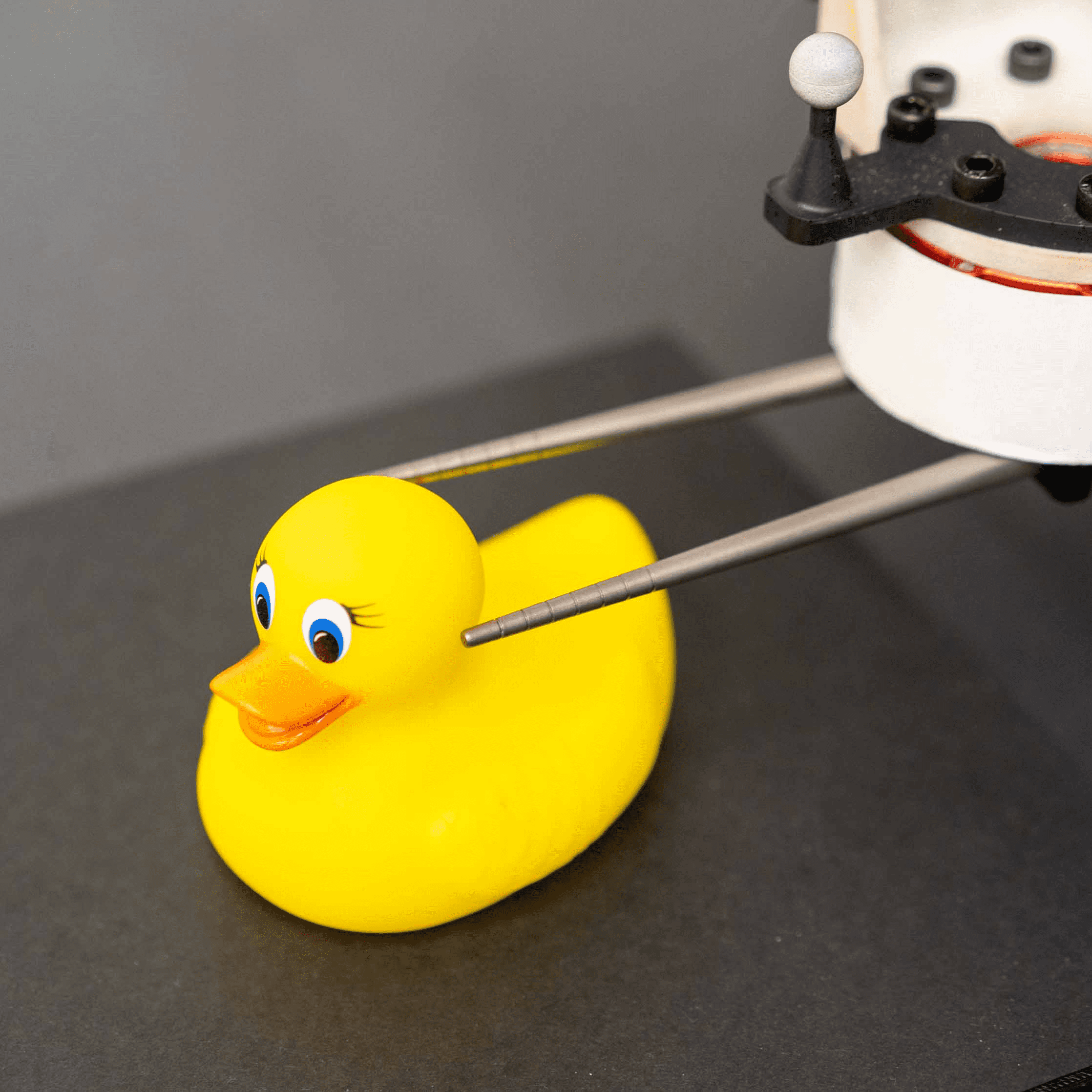}}
	\end{minipage} & Duck  & 50 (5/10) & 
           \begin{minipage}[b]{0.3\columnwidth}
		\centering
		\raisebox{-.5\height}{\includegraphics[width=\linewidth]{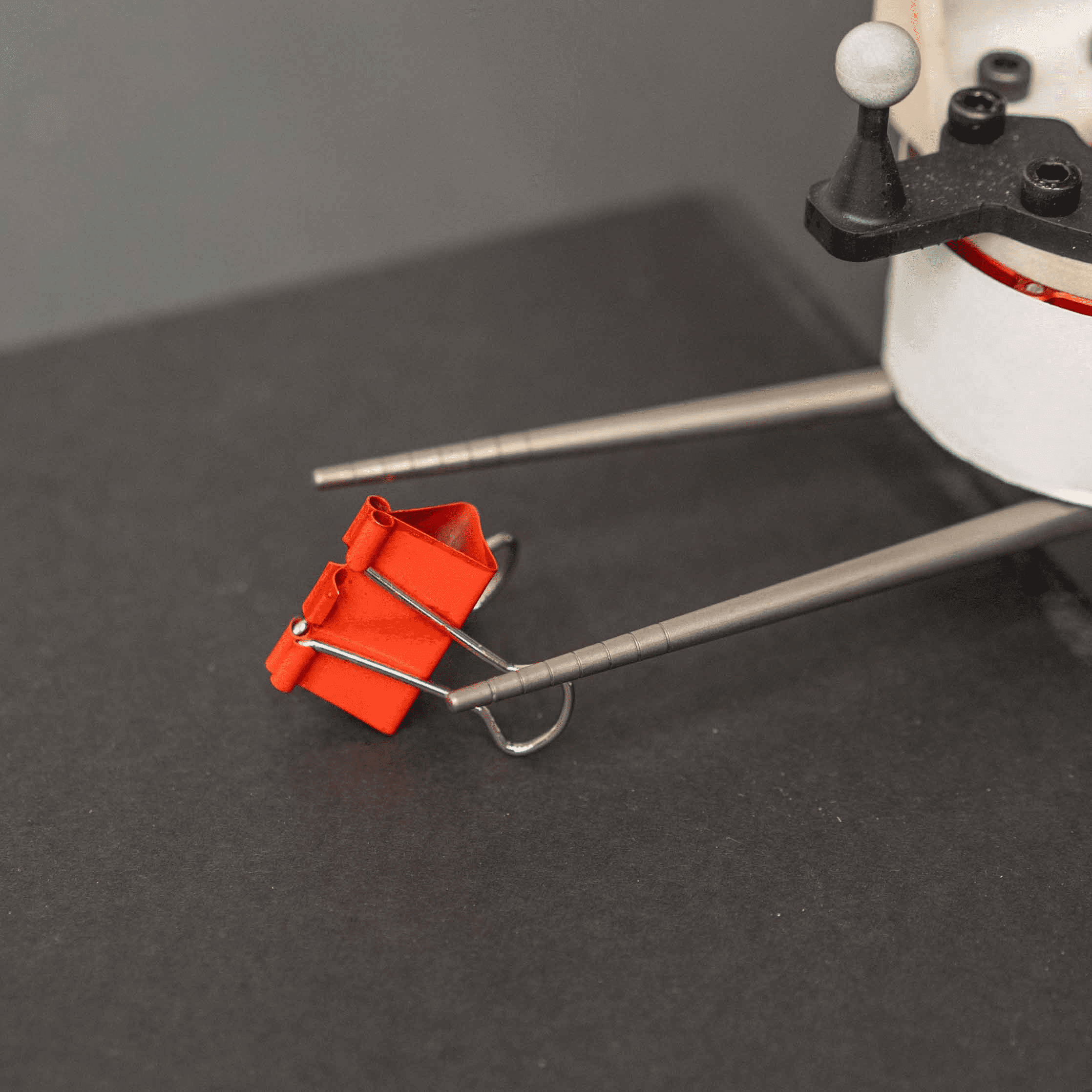}}
	\end{minipage} & Clip  & 30 (3/10) 
      \\
      \midrule
      \begin{minipage}[b]{0.3\columnwidth}
		\centering
		\raisebox{-.5\height}{\includegraphics[width=\linewidth]{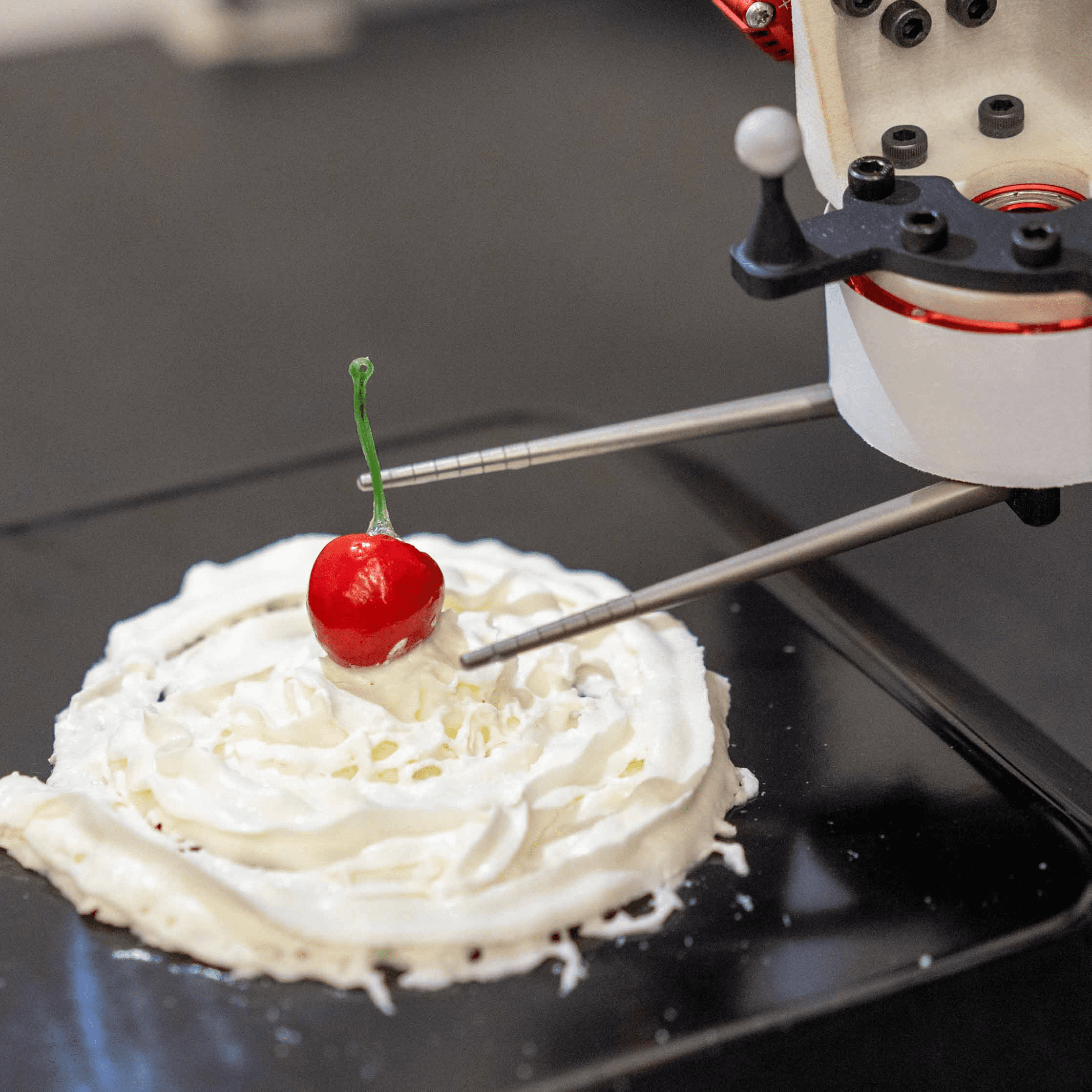}}
	\end{minipage} & Cream Cherry & 100 (10/10) &
        \begin{minipage}[b]{0.3\columnwidth}
		\centering
		\raisebox{-.5\height}{\includegraphics[width=\linewidth]{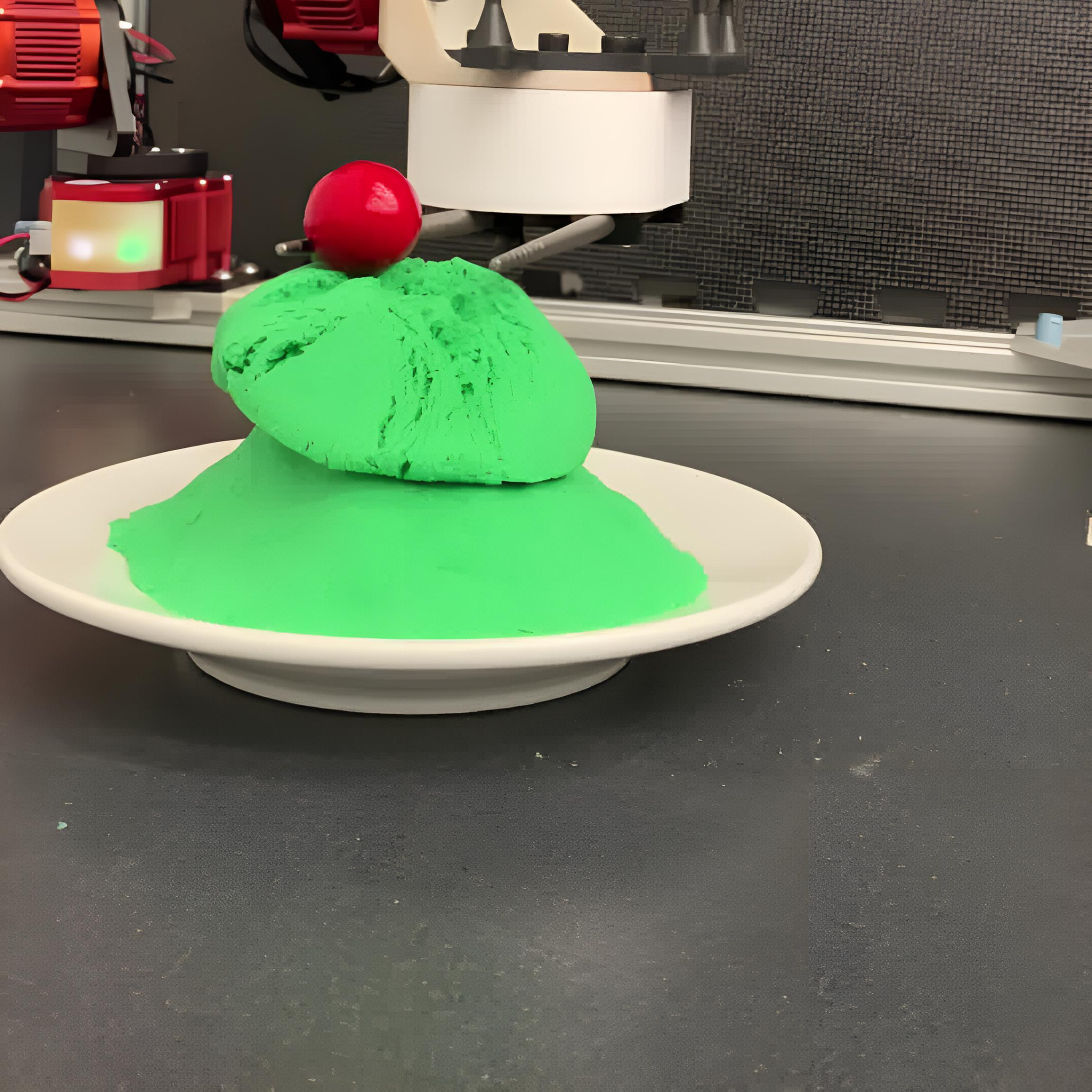}}
	\end{minipage}    & Sand Cherry  & 53.3 (8/15) 
        \\
        \midrule
              \begin{minipage}[b]{0.3\columnwidth}
		\centering
		\raisebox{-.5\height}{\includegraphics[width=\linewidth]{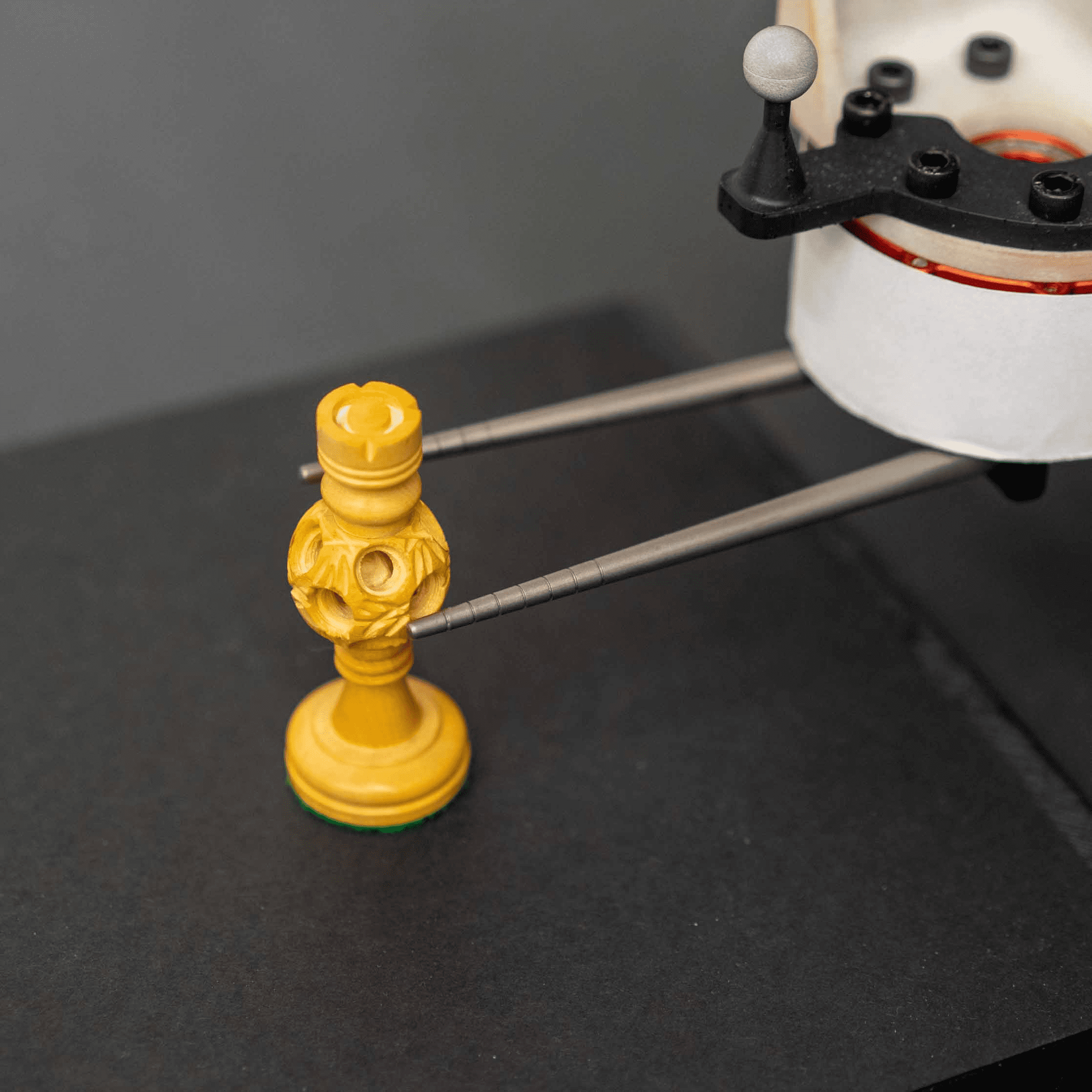}}
	\end{minipage} &  Chess  & 80 (8/10)  & 
         \begin{minipage}[b]{0.3\columnwidth}
		\centering
		\raisebox{-.5\height}{\includegraphics[width=\linewidth]{media/focus-grape.png}}
	\end{minipage} & Grapes  & 70 (7/10) 
        \\
        \midrule
 \begin{minipage}[b]{0.3\columnwidth}
		\centering
		\raisebox{-.5\height}{\includegraphics[width=\linewidth]{media/focus-tree.png}}
	\end{minipage}& Tree Cherry & 40 (4/10) &
  \begin{minipage}[b]{0.3\columnwidth}
		\centering
		\raisebox{-.5\height}{\includegraphics[width=\linewidth]{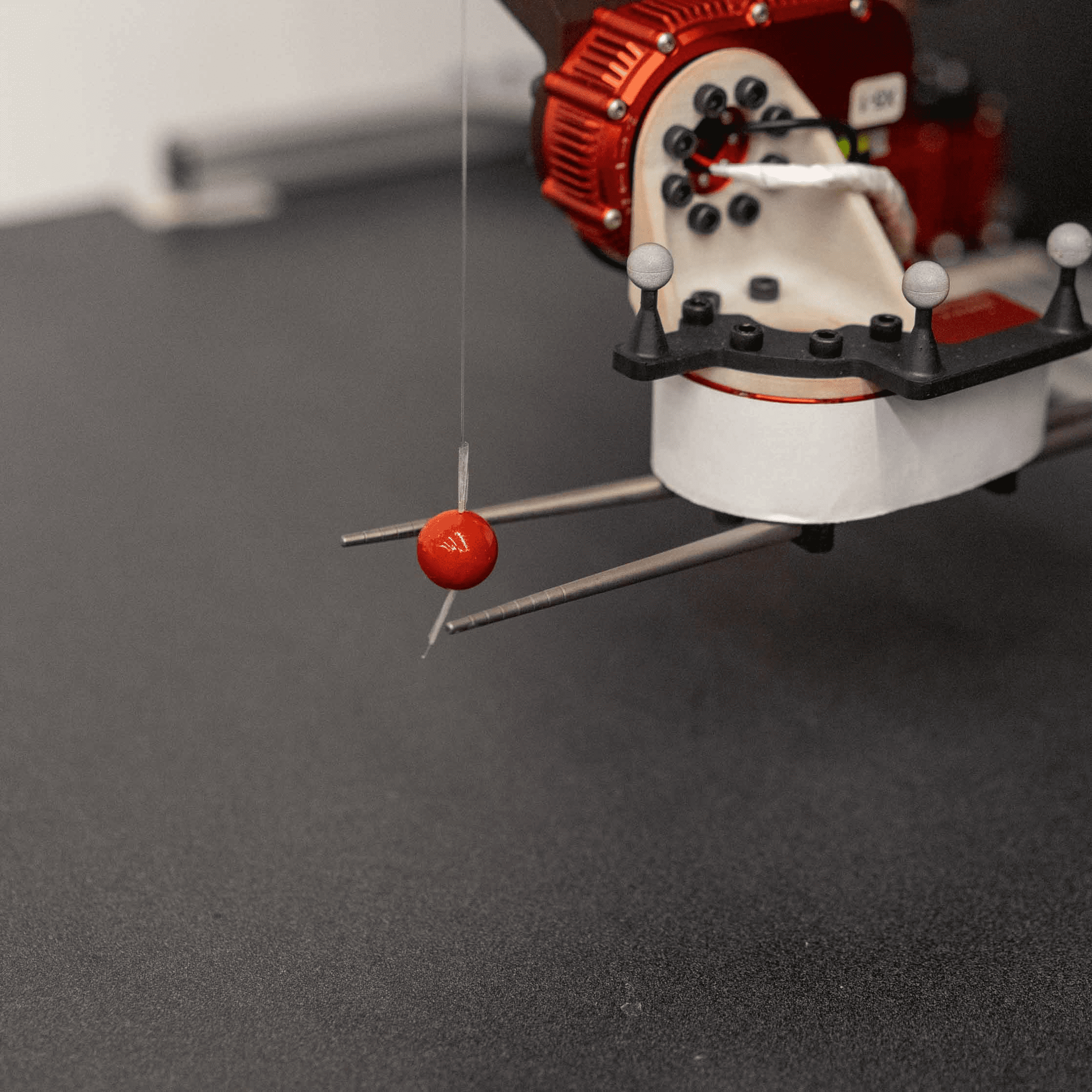}}
	\end{minipage} & Marble Ball & 100 
        \\
        \bottomrule
    \end{tabular}
    }
    \captionof{table}{Visualization and Success rates of testing tasks.}
    \label{tab:viz-real}
    % \vspace{-1em}
\end{table*}

%\paragraph{More details for the disturbance task}
%\begin{figure}[htbp]
  %\centering
%\includesvg[inkscapelatex=false,width=0.8\linewidth]{figs/disturb_rew.svg}
  %\caption{\textbf{Rewards under different disturbance }}
  %\label{fig:rl-curve-rew}
%\end{figure}

%Alongside the success rate we

Our chosen real-world evaluation tasks effectively capture our agent's ability to generalize from the proxy task to more practical settings. Particularly, these tasks feature real-world complications that are not present during training, allowing us to test the agent's ability to compensate for it at test time.

\begin{itemize}
    \item Cherry-Picking: Our agent's eponymous cherry-picking task is complicated by noisy perception affected by occlusion. The branches and leaves get in the way of the camera, resulting in biased and noisy estimates of the cherry's position. In succeeding, the agent shows its robustness to overcome these perception challenges.
    
    \item Water Cherry, Cream Cherry, Sand Cherry, and Grape: These tasks' main defining traits are the presence of new, unmodeled dynamics. In the case of the cherry floating in the water, if the cherry is disturbed it will float in the direction of the applied force. The cream cherry is relatively less mobile, but the base upon which it sits is deformable. Similarly, the cherry atop the sand is resting on a very unstable surface that is actively falling apart over time. Finally, the grape is on top of a pile of other grapes which tends to collapse when disturbed. In all four cases, there are new dynamics that are completely unrepresentative of conditions at train time. Evidently, the agent is able to generalize to these new dynamics and successfully solve the task.
    
    \item Clip, Chess, Puzzle, and Duck: These tasks showcase the agent's ability to grasp objects with completely different shapes. The agent was trained only to grasp small spheres, but in succeeding in these tasks, it shows that the agent still learns a grasping strategy that is widely applicable to other types of shapes.
\end{itemize}

\subsection{Robustness to random seeds and hyperparameters}
\label{app:sweep}

We used a standard, open-source implementation of RL algorithms, d3rlpy, to run our ablation studies in the simulation and our real-world robotic experiments. We use the default implementation of SAC and IQL in d3rlpy alongside the default hyperparameters. The only edits we make are to enable high UTD and LayerNorm. We did not change the hyperparameters or the algorithms when we ran the experiment. To enable d3rlpy to run on the real robot: we changed the location of the action normalization function in d3rlpy to ensure it will be called both for the offline dataset and online finetuning on a real robot. 

Prior to running the experiments we fixed the random seed. To verify the significance of our findings in the ablation study, we conducted seed sweeping in the simulator, with 5 consecutive random seeds $120 \sim 124$, as shown in \ref{fig:ablation-finetune}. To verify the reproducibility of our proposed system in the real world, we trained our whole system including pretraining and fine-tuning with seed $121 \sim 124$), the performance shown in \ref{fig:ablation-sim} is summarized over all random seeds ran in the real world.

% \newpage

\bibliographystyle{unsrtnat}
\bibliography{main.bib}

\begin{thebibliography}{61}
\providecommand{\natexlab}[1]{#1}
\providecommand{\url}[1]{\texttt{#1}}
\expandafter\ifx\csname urlstyle\endcsname\relax
  \providecommand{\doi}[1]{doi: #1}\else
  \providecommand{\doi}{doi: \begingroup \urlstyle{rm}\Url}\fi

\bibitem[Cutkosky(2012)]{cutkosky2012robotic}
Mark~R Cutkosky.
\newblock \emph{Robotic grasping and fine manipulation}, volume~6.
\newblock Springer Science \& Business Media, 2012.

\bibitem[Ke et~al.(2021)Ke, Wang, Bhattacharjee, Boots, and
  Srinivasa]{ke2021grasping}
Liyiming Ke, Jingqiang Wang, Tapomayukh Bhattacharjee, Byron Boots, and
  Siddhartha Srinivasa.
\newblock Grasping with chopsticks: Combating covariate shift in model-free
  imitation learning for fine manipulation.
\newblock In \emph{2021 IEEE International Conference on Robotics and
  Automation (ICRA)}, pages 6185--6191. IEEE, 2021.

\bibitem[Mason and Lynch(1993)]{mason1993dynamic}
Matthew~T Mason and Kevin~M Lynch.
\newblock Dynamic manipulation.
\newblock In \emph{Proceedings of 1993 IEEE/RSJ International Conference on
  Intelligent Robots and Systems (IROS'93)}, volume~1, pages 152--159. IEEE,
  1993.

\bibitem[Billard and Kragic(2019)]{billard2019trends}
Aude Billard and Danica Kragic.
\newblock Trends and challenges in robot manipulation.
\newblock \emph{Science}, 364\penalty0 (6446), 2019.

\bibitem[Marohn and Hanly(2004)]{marohn2004davinci}
Col Michael~R Marohn and Capt Eric~J Hanly.
\newblock Twenty-first century surgery using twenty-first century technology:
  Surgical robotics.
\newblock \emph{Current Surgery}, 61\penalty0 (5):\penalty0 466--473, 2004.

\bibitem[Yuan et~al.(2017)Yuan, Dong, and Adelson]{yuan2017gelsight}
Wenzhen Yuan, Siyuan Dong, and Edward~H Adelson.
\newblock Gelsight: High-resolution robot tactile sensors for estimating
  geometry and force.
\newblock \emph{Sensors}, 17\penalty0 (12):\penalty0 2762, 2017.

\bibitem[Bhattacharjee et~al.(2019)Bhattacharjee, Lee, Song, and
  Srinivasa]{bhattacharjee2019towards}
Tapomayukh Bhattacharjee, Gilwoo Lee, Hanjun Song, and Siddhartha~S Srinivasa.
\newblock Towards robotic feeding: Role of haptics in fork-based food
  manipulation.
\newblock \emph{IEEE Robotics and Automation Letters}, 4\penalty0 (2):\penalty0
  1485--1492, 2019.

\bibitem[Li et~al.(2019)Li, Stampfli, Xu, Malkin, Diaz, Rus, and
  Wood]{li2019vacuum}
Shuguang Li, John~J Stampfli, Helen~J Xu, Elian Malkin, Evelin~Villegas Diaz,
  Daniela Rus, and Robert~J Wood.
\newblock A vacuum-driven origami “magic-ball” soft gripper.
\newblock In \emph{2019 International Conference on Robotics and Automation
  (ICRA)}, pages 7401--7408. IEEE, 2019.

\bibitem[Zeng et~al.(2022)Zeng, Song, Yu, Donlon, Hogan, Bauza, Ma, Taylor,
  Liu, Romo, et~al.]{zeng2022robotic}
Andy Zeng, Shuran Song, Kuan-Ting Yu, Elliott Donlon, Francois~R Hogan, Maria
  Bauza, Daolin Ma, Orion Taylor, Melody Liu, Eudald Romo, et~al.
\newblock Robotic pick-and-place of novel objects in clutter with
  multi-affordance grasping and cross-domain image matching.
\newblock \emph{The International Journal of Robotics Research}, 41\penalty0
  (7):\penalty0 690--705, 2022.

\bibitem[Hwang et~al.(2022)Hwang, Ichnowski, Thananjeyan, Seita, Paradis, Fer,
  Low, and Goldberg]{hwang2022automating}
Minho Hwang, Jeffrey Ichnowski, Brijen Thananjeyan, Daniel Seita, Samuel
  Paradis, Danyal Fer, Thomas Low, and Ken Goldberg.
\newblock Automating surgical peg transfer: Calibration with deep learning can
  exceed speed, accuracy, and consistency of humans.
\newblock \emph{IEEE Transactions on Automation Science and Engineering}, 2022.

\bibitem[Lynch and Mason(1999)]{lynch1999dynamic}
Kevin~M Lynch and Matthew~T Mason.
\newblock Dynamic nonprehensile manipulation: Controllability, planning, and
  experiments.
\newblock \emph{The International Journal of Robotics Research}, 18\penalty0
  (1):\penalty0 64--92, 1999.

\bibitem[Sakurai et~al.(2016)Sakurai, Kanno, and Kawashima]{sakurai2016thin}
Haruka Sakurai, Takahiro Kanno, and Kenji Kawashima.
\newblock Thin-diameter chopsticks robot for laparoscopic surgery.
\newblock In \emph{Proceedings of the IEEE International Conference on Robotics
  and Automation (ICRA)}, pages 4122--4127, 2016.

\bibitem[Joseph et~al.(2010)Joseph, Goh, Cuevas, Donovan, Kauffman, Salas,
  Miles, Bass, and Dunkin]{joseph2010chopstick}
Rohan~A Joseph, Alvin~C Goh, Sebastian~P Cuevas, Michael~A Donovan, Matthew~G
  Kauffman, Nilson~A Salas, Brian Miles, Barbara~L Bass, and Brian~J Dunkin.
\newblock Chopstick surgery: a novel technique improves surgeon performance and
  eliminates arm collision in robotic single-incision laparoscopic surgery.
\newblock \emph{Surgical Endoscopy}, 24\penalty0 (6):\penalty0 1331--1335,
  2010.

\bibitem[Yamazaki and Masuda(2012)]{yamazaki2012autonomous}
Akira Yamazaki and Ryosuke Masuda.
\newblock Autonomous foods handling by chopsticks for meal assistant robot.
\newblock In \emph{ROBOTIK 2012; 7th German Conference on Robotics}, pages
  1--6. VDE, 2012.

\bibitem[Ramadan et~al.(2009)Ramadan, Takubo, Mae, Oohara, and
  Arai]{ramadan2009developmental}
Ahmed~A Ramadan, Tomohito Takubo, Yasushi Mae, Kenichi Oohara, and Tatsuo Arai.
\newblock Developmental process of a chopstick-like hybrid-structure
  two-fingered micromanipulator hand for 3-d manipulation of microscopic
  objects.
\newblock \emph{IEEE Transactions on Industrial Electronics}, 56\penalty0
  (4):\penalty0 1121--1135, 2009.

\bibitem[Ke et~al.(2020)Ke, Kamat, Wang, Bhattacharjee, Mavrogiannis, and
  Srinivasa]{ke2020telemanipulation}
Liyiming Ke, Ajinkya Kamat, Jingqiang Wang, Tapomayukh Bhattacharjee,
  Christoforos Mavrogiannis, and Siddhartha~S Srinivasa.
\newblock Telemanipulation with chopsticks: Analyzing human factors in user
  demonstrations.
\newblock In \emph{2020 IEEE/RSJ International Conference on Intelligent Robots
  and Systems (IROS)}, pages 11539--11546. IEEE, 2020.

\bibitem[Mason et~al.(2011)Mason, Srinivasa, and Vazquez]{mason2011generality}
Matthew~T Mason, Siddhartha~S Srinivasa, and Andres~S Vazquez.
\newblock Generality and simple hands.
\newblock In \emph{Robotics Research}, pages 345--361. Springer, 2011.

\bibitem[Chang et~al.(2007)Chang, Huang, Chen, and Wang]{chang2007pincer}
Bao-Chi Chang, Biing-Shiun Huang, Ching-Kong Chen, and Shyh-Jen Wang.
\newblock The pincer chopsticks: The investigation of a new utensil in pinching
  function.
\newblock \emph{Applied ergonomics}, 38\penalty0 (3):\penalty0 385--390, 2007.

\bibitem[Mordatch et~al.(2012)Mordatch, Todorov, and
  Popovi{\'c}]{mordatch2012discovery}
Igor Mordatch, Emanuel Todorov, and Zoran Popovi{\'c}.
\newblock Discovery of complex behaviors through contact-invariant
  optimization.
\newblock \emph{ACM Transactions on Graphics (ToG)}, 31\penalty0 (4):\penalty0
  1--8, 2012.

\bibitem[Kumar et~al.(2016)Kumar, Todorov, and Levine]{kumar2016optimal}
Vikash Kumar, Emanuel Todorov, and Sergey Levine.
\newblock Optimal control with learned local models: Application to dexterous
  manipulation.
\newblock In \emph{2016 IEEE International Conference on Robotics and
  Automation (ICRA)}, pages 378--383. IEEE, 2016.

\bibitem[Hogan and Rodriguez(2020)]{hogan2020reactive}
Francois~R Hogan and Alberto Rodriguez.
\newblock Reactive planar non-prehensile manipulation with hybrid model
  predictive control.
\newblock \emph{The International Journal of Robotics Research}, 39\penalty0
  (7):\penalty0 755--773, 2020.

\bibitem[Schaal(2006)]{schaal2006dynamic}
Stefan Schaal.
\newblock Dynamic movement primitives-a framework for motor control in humans
  and humanoid robotics.
\newblock \emph{Adaptive motion of animals and machines}, pages 261--280, 2006.

\bibitem[Kalashnikov et~al.(2018)Kalashnikov, Irpan, Pastor, Ibarz, Herzog,
  Jang, Quillen, Holly, Kalakrishnan, Vanhoucke, et~al.]{kalashnikov2018qt}
Dmitry Kalashnikov, Alex Irpan, Peter Pastor, Julian Ibarz, Alexander Herzog,
  Eric Jang, Deirdre Quillen, Ethan Holly, Mrinal Kalakrishnan, Vincent
  Vanhoucke, et~al.
\newblock Qt-opt: Scalable deep reinforcement learning for vision-based robotic
  manipulation.
\newblock \emph{arXiv preprint arXiv:1806.10293}, 2018.

\bibitem[Zhu et~al.(2019)Zhu, Gupta, Rajeswaran, Levine, and
  Kumar]{zhu2019dexterous}
Henry Zhu, Abhishek Gupta, Aravind Rajeswaran, Sergey Levine, and Vikash Kumar.
\newblock Dexterous manipulation with deep reinforcement learning: Efficient,
  general, and low-cost.
\newblock In \emph{2019 International Conference on Robotics and Automation
  (ICRA)}, pages 3651--3657. IEEE, 2019.

\bibitem[Zhu et~al.(2020)Zhu, Yu, Gupta, Shah, Hartikainen, Singh, Kumar, and
  Levine]{zhu20ingredients}
Henry Zhu, Justin Yu, Abhishek Gupta, Dhruv Shah, Kristian Hartikainen, Avi
  Singh, Vikash Kumar, and Sergey Levine.
\newblock The ingredients of real world robotic reinforcement learning.
\newblock In \emph{8th International Conference on Learning Representations,
  {ICLR} 2020, Addis Ababa, Ethiopia, April 26-30, 2020}. OpenReview.net, 2020.
\newblock URL \url{https://openreview.net/forum?id=rJe2syrtvS}.

\bibitem[Peng et~al.(2020)Peng, Coumans, Zhang, Lee, Tan, and
  Levine]{peng2020learning}
Xue~Bin Peng, Erwin Coumans, Tingnan Zhang, Tsang-Wei Lee, Jie Tan, and Sergey
  Levine.
\newblock Learning agile robotic locomotion skills by imitating animals.
\newblock \emph{arXiv preprint arXiv:2004.00784}, 2020.

\bibitem[Mahler et~al.(2016)Mahler, Pokorny, Hou, Roderick, Laskey, Aubry,
  Kohlhoff, Kr{\"o}ger, Kuffner, and Goldberg]{mahler2016dex}
Jeffrey Mahler, Florian~T Pokorny, Brian Hou, Melrose Roderick, Michael Laskey,
  Mathieu Aubry, Kai Kohlhoff, Torsten Kr{\"o}ger, James Kuffner, and Ken
  Goldberg.
\newblock Dex-net 1.0: A cloud-based network of 3d objects for robust grasp
  planning using a multi-armed bandit model with correlated rewards.
\newblock In \emph{IEEE International Conference on Robotics and Automation
  (ICRA)}, pages 1957--1964. IEEE, 2016.

\bibitem[Zeng et~al.(2021)Zeng, Florence, Tompson, Welker, Chien, Attarian,
  Armstrong, Krasin, Duong, Sindhwani, et~al.]{zeng2021transporter}
Andy Zeng, Pete Florence, Jonathan Tompson, Stefan Welker, Jonathan Chien,
  Maria Attarian, Travis Armstrong, Ivan Krasin, Dan Duong, Vikas Sindhwani,
  et~al.
\newblock Transporter networks: Rearranging the visual world for robotic
  manipulation.
\newblock In \emph{Conference on Robot Learning}, pages 726--747. PMLR, 2021.

\bibitem[Calli et~al.(2017)Calli, Singh, Bruce, Walsman, Konolige, Srinivasa,
  Abbeel, and Dollar]{calli2017yale}
Berk Calli, Arjun Singh, James Bruce, Aaron Walsman, Kurt Konolige, Siddhartha
  Srinivasa, Pieter Abbeel, and Aaron~M Dollar.
\newblock Yale-{CMU-Berkeley} dataset for robotic manipulation research.
\newblock \emph{The International Journal of Robotics Research}, 36\penalty0
  (3):\penalty0 261--268, 2017.

\bibitem[Hashimoto et~al.(2001)Hashimoto, Namiki, and
  Ishikawa]{hashimoto2001visuomotor}
Koichi Hashimoto, Akio Namiki, and Masatoshi Ishikawa.
\newblock A visuomotor control architecture for high-speed grasping.
\newblock In \emph{Proceedings of the 40th IEEE Conference on Decision and
  Control (Cat. No. 01CH37228)}, volume~1, pages 15--20. IEEE, 2001.

\bibitem[Haarnoja et~al.(2018{\natexlab{a}})Haarnoja, Ha, Zhou, Tan, Tucker,
  and Levine]{haarnoja2018learning}
Tuomas Haarnoja, Sehoon Ha, Aurick Zhou, Jie Tan, George Tucker, and Sergey
  Levine.
\newblock Learning to walk via deep reinforcement learning.
\newblock \emph{arXiv preprint arXiv:1812.11103}, 2018{\natexlab{a}}.

\bibitem[Haarnoja et~al.(2018{\natexlab{b}})Haarnoja, Zhou, Hartikainen,
  Tucker, Ha, Tan, Kumar, Zhu, Gupta, Abbeel, et~al.]{haarnoja2018soft}
Tuomas Haarnoja, Aurick Zhou, Kristian Hartikainen, George Tucker, Sehoon Ha,
  Jie Tan, Vikash Kumar, Henry Zhu, Abhishek Gupta, Pieter Abbeel, et~al.
\newblock Soft actor-critic algorithms and applications.
\newblock \emph{arXiv preprint arXiv:1812.05905}, 2018{\natexlab{b}}.

\bibitem[W{\"u}thrich et~al.(2020)W{\"u}thrich, Widmaier, Grimminger, Akpo,
  Joshi, Agrawal, Hammoud, Khadiv, Bogdanovic, Berenz,
  et~al.]{wuthrich2020trifinger}
Manuel W{\"u}thrich, Felix Widmaier, Felix Grimminger, Joel Akpo, Shruti Joshi,
  Vaibhav Agrawal, Bilal Hammoud, Majid Khadiv, Miroslav Bogdanovic, Vincent
  Berenz, et~al.
\newblock Trifinger: An open-source robot for learning dexterity.
\newblock \emph{arXiv preprint arXiv:2008.03596}, 2020.

\bibitem[Rajeswaran et~al.(2017)Rajeswaran, Kumar, Gupta, Vezzani, Schulman,
  Todorov, and Levine]{rajeswaran2017learning}
Aravind Rajeswaran, Vikash Kumar, Abhishek Gupta, Giulia Vezzani, John
  Schulman, Emanuel Todorov, and Sergey Levine.
\newblock Learning complex dexterous manipulation with deep reinforcement
  learning and demonstrations.
\newblock \emph{arXiv preprint arXiv:1709.10087}, 2017.

\bibitem[Chen et~al.(2021)Chen, Wang, Zhou, and Ross]{UTD}
Xinyue Chen, Che Wang, Zijian Zhou, and Keith Ross.
\newblock Randomized ensembled double q-learning: Learning fast without a
  model.
\newblock \emph{arXiv preprint arXiv:2101.05982}, 2021.

\bibitem[Smith et~al.(2022)Smith, Kostrikov, and Levine]{smith2022walk}
Laura Smith, Ilya Kostrikov, and Sergey Levine.
\newblock A walk in the park: Learning to walk in 20 minutes with model-free
  reinforcement learning.
\newblock \emph{arXiv preprint arXiv:2208.07860}, 2022.

\bibitem[Lange et~al.(2012)Lange, Gabel, and Riedmiller]{lange2012batch}
Sascha Lange, Thomas Gabel, and Martin Riedmiller.
\newblock Batch reinforcement learning.
\newblock \emph{Reinforcement learning: State-of-the-art}, pages 45--73, 2012.

\bibitem[Levine et~al.(2020)Levine, Kumar, Tucker, and Fu]{levine2020offline}
Sergey Levine, Aviral Kumar, George Tucker, and Justin Fu.
\newblock Offline reinforcement learning: Tutorial, review, and perspectives on
  open problems.
\newblock \emph{arXiv preprint arXiv:2005.01643}, 2020.

\bibitem[Peng et~al.(2019)Peng, Kumar, Zhang, and Levine]{awr}
Xue~Bin Peng, Aviral Kumar, Grace Zhang, and Sergey Levine.
\newblock Advantage-weighted regression: Simple and scalable off-policy
  reinforcement learning.
\newblock \emph{arXiv preprint arXiv:1910.00177}, 2019.

\bibitem[Fujimoto et~al.(2018)Fujimoto, Meger, and Precup]{fujimoto2018off}
Scott Fujimoto, David Meger, and Doina Precup.
\newblock Off-policy deep reinforcement learning without exploration. corr
  abs/1812.02900 (2018).
\newblock \emph{arXiv preprint arXiv:1812.02900}, 2018.

\bibitem[Kidambi et~al.(2020)Kidambi, Rajeswaran, Netrapalli, and
  Joachims]{kidambi2020morel}
Rahul Kidambi, Aravind Rajeswaran, Praneeth Netrapalli, and Thorsten Joachims.
\newblock Morel: Model-based offline reinforcement learning.
\newblock \emph{Advances in neural information processing systems},
  33:\penalty0 21810--21823, 2020.

\bibitem[Kumar et~al.(2020)Kumar, Zhou, Tucker, and
  Levine]{kumar2020conservative}
Aviral Kumar, Aurick Zhou, George Tucker, and Sergey Levine.
\newblock Conservative q-learning for offline reinforcement learning.
\newblock \emph{Advances in Neural Information Processing Systems},
  33:\penalty0 1179--1191, 2020.

\bibitem[Kostrikov et~al.(2021)Kostrikov, Nair, and
  Levine]{kostrikov2021offline}
Ilya Kostrikov, Ashvin Nair, and Sergey Levine.
\newblock Offline reinforcement learning with implicit q-learning.
\newblock \emph{arXiv preprint arXiv:2110.06169}, 2021.

\bibitem[Zhou et~al.(2022)Zhou, Ke, Srinivasa, Gupta, Rajeswaran, and
  Kumar]{zhou2022real}
Gaoyue Zhou, Liyiming Ke, Siddhartha Srinivasa, Abhinav Gupta, Aravind
  Rajeswaran, and Vikash Kumar.
\newblock Real world offline reinforcement learning with realistic data source.
\newblock \emph{arXiv preprint arXiv:2210.06479}, 2022.

\bibitem[Hong et~al.(2022)Hong, Kumar, and Levine]{hong2022confidence}
Joey Hong, Aviral Kumar, and Sergey Levine.
\newblock Confidence-conditioned value functions for offline reinforcement
  learning.
\newblock \emph{arXiv preprint arXiv:2212.04607}, 2022.

\bibitem[Chaumette and Hutchinson(2006)]{chaumette2006visual}
Fran{\c{c}}ois Chaumette and Seth Hutchinson.
\newblock Visual servo control. {I. Basic approaches}.
\newblock \emph{IEEE Robotics \& Automation Magazine}, 13\penalty0
  (4):\penalty0 82--90, 2006.

\bibitem[Hutchinson et~al.(1996)Hutchinson, Hager, and
  Corke]{hutchinson1996tutorial}
Seth Hutchinson, Gregory~D Hager, and Peter~I Corke.
\newblock A tutorial on visual servo control.
\newblock \emph{IEEE transactions on robotics and automation}, 12\penalty0
  (5):\penalty0 651--670, 1996.

\bibitem[Sanderson and Weiss(1983)]{sanderson1983adaptive}
Arthur~C Sanderson and Lee~E Weiss.
\newblock Adaptive visual servo control of robots.
\newblock \emph{Robot vision}, pages 107--116, 1983.

\bibitem[Garrett et~al.(2021)Garrett, Chitnis, Holladay, Kim, Silver,
  Kaelbling, and Lozano-P{\'e}rez]{garrett2021integrated}
Caelan~Reed Garrett, Rohan Chitnis, Rachel Holladay, Beomjoon Kim, Tom Silver,
  Leslie~Pack Kaelbling, and Tom{\'a}s Lozano-P{\'e}rez.
\newblock Integrated task and motion planning.
\newblock \emph{Annual review of control, robotics, and autonomous systems},
  4:\penalty0 265--293, 2021.

\bibitem[Zhang et~al.(2022)Zhang, Zhu, Ding, Zhu, Stone, and
  Zhang]{zhang2022visually}
Xiaohan Zhang, Yifeng Zhu, Yan Ding, Yuke Zhu, Peter Stone, and Shiqi Zhang.
\newblock Visually grounded task and motion planning for mobile manipulation.
\newblock In \emph{2022 International Conference on Robotics and Automation
  (ICRA)}, pages 1925--1931. IEEE, 2022.

\bibitem[Chi et~al.(2022)Chi, Burchfiel, Cousineau, Feng, and
  Song]{chi2022iterative}
Cheng Chi, Benjamin Burchfiel, Eric Cousineau, Siyuan Feng, and Shuran Song.
\newblock Iterative residual policy: for goal-conditioned dynamic manipulation
  of deformable objects.
\newblock \emph{arXiv preprint arXiv:2203.00663}, 2022.

\bibitem[Okumura et~al.(2011)Okumura, Oku, and Ishikawa]{okumura2011high}
Kohei Okumura, Hiromasa Oku, and Masatoshi Ishikawa.
\newblock High-speed gaze controller for millisecond-order pan/tilt camera.
\newblock In \emph{2011 IEEE International Conference on Robotics and
  Automation}, pages 6186--6191. IEEE, 2011.

\bibitem[Peng et~al.(2018)Peng, Andrychowicz, Zaremba, and Abbeel]{peng2018sim}
Xue~Bin Peng, Marcin Andrychowicz, Wojciech Zaremba, and Pieter Abbeel.
\newblock Sim-to-real transfer of robotic control with dynamics randomization.
\newblock In \emph{2018 IEEE international conference on robotics and
  automation (ICRA)}, pages 3803--3810. IEEE, 2018.

\bibitem[Ba et~al.(2016)Ba, Kiros, and Hinton]{layernorm}
Jimmy~Lei Ba, Jamie~Ryan Kiros, and Geoffrey~E Hinton.
\newblock Layer normalization.
\newblock \emph{arXiv preprint arXiv:1607.06450}, 2016.

\bibitem[Todorov et~al.(2012)Todorov, Erez, and Tassa]{todorov12mujoco}
Emanuel Todorov, Tom Erez, and Yuval Tassa.
\newblock Mujoco: {A} physics engine for model-based control.
\newblock In \emph{2012 {IEEE/RSJ} International Conference on Intelligent
  Robots and Systems, {IROS} 2012, Vilamoura, Algarve, Portugal, October 7-12,
  2012}, pages 5026--5033. {IEEE}, 2012.
\newblock \doi{10.1109/IROS.2012.6386109}.
\newblock URL \url{https://doi.org/10.1109/IROS.2012.6386109}.

\bibitem[Tedrake(2023)]{underactuated}
Russ Tedrake.
\newblock \emph{Underactuated Robotics}.
\newblock 2023.
\newblock URL \url{https://underactuated.csail.mit.edu}.

\bibitem[Pomerleau(1988)]{pomerleau1988alvinn}
Dean~A Pomerleau.
\newblock Alvinn: An autonomous land vehicle in a neural network.
\newblock \emph{Advances in neural information processing systems}, 1, 1988.

\bibitem[Seno and Imai(2022)]{d3rlpy}
Takuma Seno and Michita Imai.
\newblock d3rlpy: An offline deep reinforcement learning library.
\newblock \emph{Journal of Machine Learning Research}, 23\penalty0
  (315):\penalty0 1--20, 2022.
\newblock URL \url{http://jmlr.org/papers/v23/22-0017.html}.

\bibitem[Summers et~al.(2020)Summers, Lowrey, Rajeswaran, Srinivasa, and
  Todorov]{summers2020lyceum}
Colin Summers, Kendall Lowrey, Aravind Rajeswaran, Siddhartha Srinivasa, and
  Emanuel Todorov.
\newblock Lyceum: An efficient and scalable ecosystem for robot learning.
\newblock In \emph{Learning for Dynamics and Control}, pages 793--803. PMLR,
  2020.

\bibitem[Riedmiller et~al.(2018)Riedmiller, Hafner, Lampe, Neunert, Degrave,
  Wiele, Mnih, Heess, and Springenberg]{riedmiller2018learning}
Martin Riedmiller, Roland Hafner, Thomas Lampe, Michael Neunert, Jonas Degrave,
  Tom Wiele, Vlad Mnih, Nicolas Heess, and Jost~Tobias Springenberg.
\newblock Learning by playing solving sparse reward tasks from scratch.
\newblock In \emph{International conference on machine learning}, pages
  4344--4353. PMLR, 2018.

\bibitem[Bochkovskiy et~al.(2020)Bochkovskiy, Wang, and
  Liao]{bochkovskiy2020yolov4}
Alexey Bochkovskiy, Chien-Yao Wang, and Hong-Yuan~Mark Liao.
\newblock Yolov4: Optimal speed and accuracy of object detection.
\newblock \emph{arXiv preprint arXiv:2004.10934}, 2020.

\end{thebibliography}

\end{document}